\documentclass[10pt,twocolumn,letterpaper]{article}

\usepackage[pagenumbers]{cvpr} 

\usepackage{times}
\usepackage{epsfig}
\usepackage{graphicx}
\usepackage{amsmath}
\usepackage{amssymb}

\usepackage{booktabs}
\usepackage{multirow}
\usepackage{graphicx}
\usepackage{amsmath}
\usepackage{bbm}
\usepackage{booktabs}
\usepackage{placeins}
\usepackage[dvipsnames]{xcolor}
\usepackage{algorithm}
\usepackage[noend]{algpseudocode}
\usepackage{wrapfig}
\usepackage{colortbl}
\usepackage{overpic}
\usepackage{pifont}
\newcommand{\cmark}{\ding{51}}%
\newcommand{\xmark}{\ding{55}}%

\newcommand{\cutcaptionup}{\vspace*{-6pt}}
\newcommand{\cutcaptiondown}{\vspace*{-2pt}}

\newcommand*\samethanks[1][\value{footnote}]{\footnotemark[#1]}
\newcommand{\fullmodelname}{Multi-modality Scene Tokenization}
\newcommand{\modelname}{MoST}
\newcommand{\datasetname}{WOMD}
\newcommand{\plh}{%
  {\ooalign{$\phantom{0}$\cr\hidewidth$\scriptstyle\times$\cr}}%
}

\definecolor{cvprblue}{rgb}{0.21,0.49,0.74}
\usepackage[pagebackref,breaklinks,colorlinks,citecolor=cvprblue]{hyperref}

\usepackage[capitalize]{cleveref}
\crefname{section}{Sec.}{Secs.}
\Crefname{section}{Section}{Sections}
\Crefname{table}{Table}{Tables}
\crefname{table}{Tab.}{Tabs.}

\title{MoST: \underline{M}ulti-m\underline{o}dality \underline{S}cene \underline{T}okenization for Motion Prediction}

\author{Norman Mu\thanks{Equal contribution} \quad
Jingwei Ji\samethanks \quad
Zhenpei Yang\samethanks \quad
Nate Harada\samethanks \quad
Haotian Tang\samethanks \quad
Kan Chen\samethanks \\
Charles R. Qi \quad
Runzhou Ge \quad
Kratarth Goel \quad
Zoey Yang \quad
Scott Ettinger \\
Rami Al-Rfou \quad
Dragomir Anguelov \quad 
Yin Zhou\thanks{Corresponding author}
\\ Waymo LLC}

\begin{document}
\maketitle
\begin{abstract}
Many existing motion prediction approaches rely on symbolic perception outputs to generate agent trajectories, such as bounding boxes, road graph information and traffic lights. This symbolic representation is a high-level abstraction of the real world, which may render the motion prediction model vulnerable to perception errors (e.g., failures in detecting open-vocabulary obstacles) while missing salient information from the scene context (e.g., poor road conditions). An alternative paradigm is end-to-end learning from raw sensors. However, this approach suffers from the lack of interpretability and requires significantly more training resources. In this work, we propose tokenizing the visual world into a compact set of scene elements and then leveraging  pre-trained image foundation models and LiDAR neural networks to encode all the scene elements in an open-vocabulary manner. The image foundation model enables our scene tokens to encode the general knowledge of the open world while the LiDAR neural network encodes geometry information. Our proposed representation can efficiently encode the multi-frame multi-modality observations with a few hundred tokens and is compatible with most transformer-based architectures. To evaluate our method, we have augmented Waymo Open Motion Dataset with camera embeddings. Experiments over Waymo Open Motion Dataset show that our approach leads to significant performance improvements over the state-of-the-art.

\end{abstract}

\begin{figure}[ht]
    \centering
    \includegraphics[width=\linewidth]{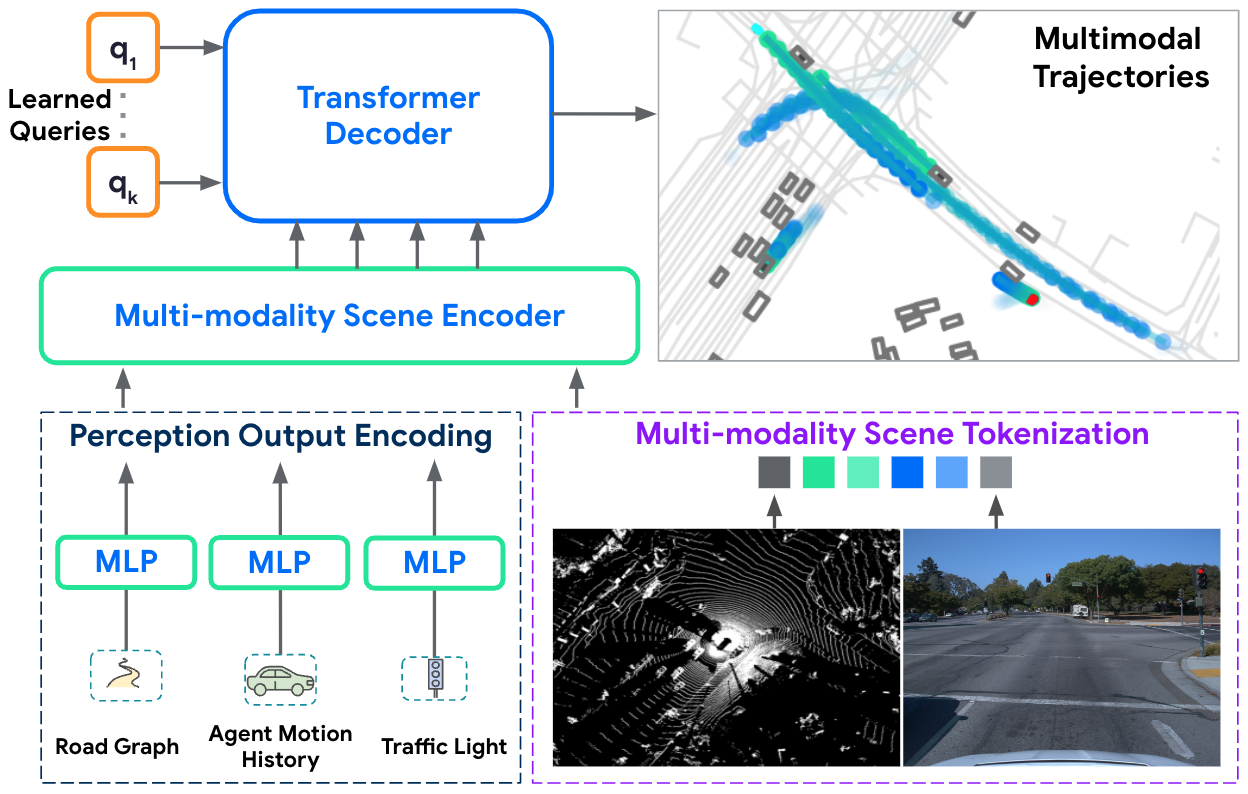}
    \cutcaptionup
    \cutcaptionup
    \cutcaptionup
    \caption{Overview of the proposed motion prediction paradigm. It fuses symbolic perception output and our multi-modality scene tokens. While symbolic representation offers a convenient world abstraction, the multi-modality scene tokens links behavior models directly to sensor observations via token embeddings.}
    \label{fig:teaser}
    \vspace{-1em}
\end{figure}    
\section{Introduction}
\label{sec:intro}

In order to safely and effectively operate in complex environments, autonomous systems must model the behavior of nearby agents.
These motion prediction models now often rely on symbolic perception outputs such as 3D bounding box tracks to represent agent states, rather than directly processing sensor inputs.
These representations reduce input dimensionality, facilitating computationally efficient model training.
Additionally, since inputs such as 3D boxes are easily rearranged and manipulated, it is possible to construct many hypothetical scenarios leading to efficient simulation and testing.
Yet in order to continue improving the accuracy and robustness of behavior models, it may be necessary to feed the models higher-fidelity sensor features.
For instance, pedestrian pose and gaze offer richer cues than mere bounding boxes for motion prediction. Moreover, many scene elements like lane markings cannot be well represented by boxes. 
Furthermore, scene context (e.g., road surface conditions, hazardous locations) is difficult to characterize with symbolic representations.  
Manually crafting representation for diverse concepts demands considerable engineering effort in implementation, training, and evaluation. Instead, we want the behavior model to directly access the raw sensor data and determine what and how to encode.

Deep learning models' performance generally improves when we replace hand-crafted features, designed to encode inductive bias according to expert domain knowledge, with the directly observed feature as long we scale compute and data accordingly.
But learning to predict complex patterns such as agent behavior directly from very high-dimensional sensor inputs (e.g. many high-resolution LiDAR and camera sensors all operating at high frequency) is an extremely challenging learning problem.
It requires learning to organize many hundreds of thousands of points and pixels across time into meaningful representations.
Moreover, the intermediate representations of fully end-to-end systems are far more difficult to validate and inspect.

Rather than choosing strictly between the two approaches, we instead propose combining existing symbolic representations with learned tokens encoding scene information.
We first decompose the scene into a compact set of disjoint elements representing ground regions, perception-detected agents and open-set objects, based on ground plane fitting and connected component analysis. We then leverage large pre-trained 2D image models and 3D point cloud models to encode these scene elements into ``tokens''. The 2D image models are trained on Internet-scale data, and show impressive capabilities in understanding the open visual world. These tokens encapsulate relevant information for reasoning about the environment, such as object semantics, object geometry as well as the scene context. 
We compactly represent multi-modality information about ground, agents and open-set objects into a few hundred tokens, which we later feed to Wayformer-like network~\cite{nayakanti2023wayformer} alongside tokens encoding agent position and velocity, road graph, and traffic signals.
All tokens are processed via a linear projection into same dimension and self-attention layers.

To evaluate our method, we introduce camera embeddings to the Waymo Open Motion Dataset (WOMD)~\cite{Ettinger_2021_ICCV}. With LiDAR points~\cite{chen2023womd} and camera embeddings, WOMD has become a large-scale multi-modal dataset for motion prediction.
On the \datasetname, our model, which combines learned and symbolic scene tokens, brings 6.6\% relative improvement on soft mAP or 10.3\% relative improvement on minADE.
While we obtain the strongest results with the recently released image backbone from~\cite{kirillov2023segment}, other pre-trained image models ~\cite{radford2021learning, oquab2023dinov2} also yield considerable gains.
We further analyze the performance of our trajectory prediction model under challenging scenarios.
Notably, we discover that even in the presence of imperfect symbolic perception outputs and incomplete road graph information, our model maintains exceptional robustness and accuracy.

Our contributions are three-fold:
\begin{itemize}
    \item We have augmented WOMD into a large-scale multi-modal dataset to support research in end-to-end learning. Camera embeddings are released to the community.
    \item We have conducted a thorough study of modeling ideas of varying complexity to demonstrate the value of those sensory inputs in motion prediction.
    \item We have proposed a novel method MoST, which effectively leverages the multi-modality data and leads to significant performance improvement.  
\end{itemize}

\section{Related Works}
\label{sec:related_works}

\begin{figure*}
\centering
\includegraphics[width=0.95\linewidth]{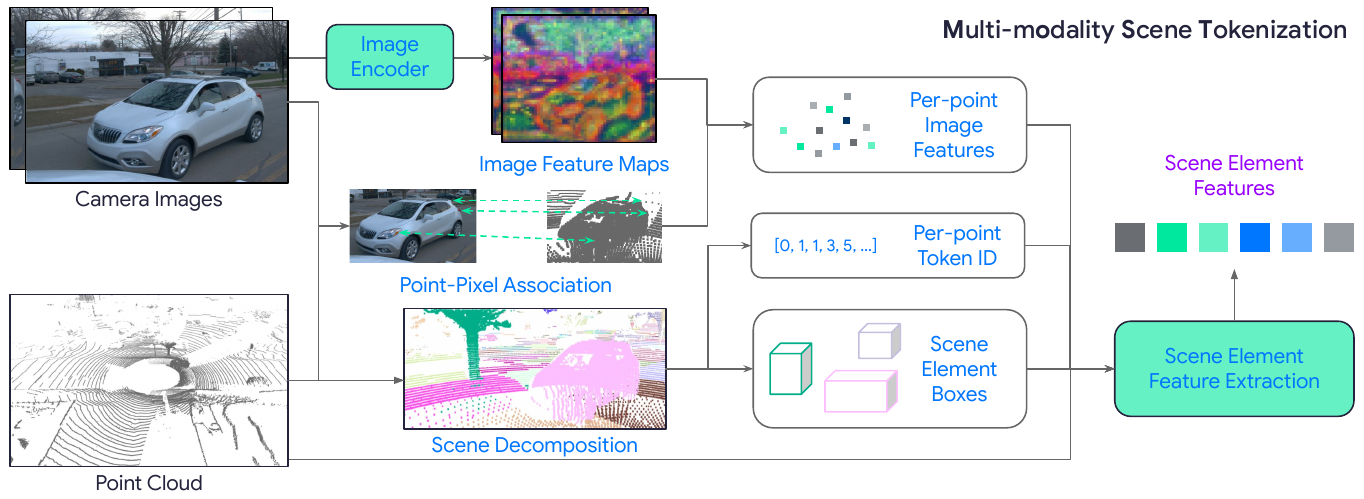}

\cutcaptionup

\caption{Overview of the proposed  Multi-modality Scene Tokenization. Our method takes as input multi-view camera images and a full scene point cloud. We leverage a pre-trained image foundation model to obtain descriptive feature maps and decompose the scene into disjoint elements via clustering. Based on the sensor calibration information between camera and LiDAR, we obtain point-wise image features. From scene decomposition, we assign each point with a token/cluster id and derive box information for each element. Finally, we extract one feature embedding for each scene element.}

\cutcaptiondown
\label{fig:flowchart}
\end{figure*}

\paragraph{Motion Prediction for Autonomous Driving} The increasing interest in autonomous driving has led to a significant focus on motion prediction~\cite{multipath_2019,DenseTNT_Gu_2021_ICCV,MultiPathPP_2021,nayakanti2023wayformer,shi2022motion,seff2023motionlm}. Early methods~\cite{park2020diverse,marchetti2020mantra,chai2019multipath,casas2018intentnet,cui2019multimodal,hong2019rules,biktairov2020prank,casas2021mp3,gilles2021home} rasterize the input scene into a 2D image, followed by processing using convolutional neural networks (CNNs). However, as a result of the inherent lossiness in the rasterization process, contemporary research has shifted its focus towards representing road elements, such as object bounding boxes, road graphs, and traffic light signals, as discrete graph nodes~\cite{gao2020vectornet}. These elements are then directly processed using graph neural networks (GNNs)~\cite{casas2020spagnn,liang2020learning,khandelwal2020if,gilles2022gohome}. Another stream of research also employs this discrete set representation for scene elements but processes them using recurrent neural networks~\cite{gupta2018social,tang2019multiple,marchetti2020mantra,MultiPathPP_2021,buhet2020plop,salzmann2020trajectron++}, rather than GNNs. Thanks to the rapid advancement of transformer-based architectures in natural language processing and computer vision, the latest state-of-the-art motion predictors also extensively incorporate the attention mechanism~\cite{nayakanti2023wayformer,ngiam2021scene,shi2022motion,jia2023hdgt,seff2023motionlm}. More recently, the community has also started to study interactive behavior prediction, which jointly models the future motion of multiple objects~\cite{luo2023jfp,tolstaya2021identifying,song2020pip,sun2022m2i}.

\paragraph{End-to-end Autonomous Driving} The concept of end-to-end learning-based autonomous driving systems started in the late 1980s~\cite{pomerleau1988alvinn}. Since then, researchers have developed differentiable modules that connect perception and behavior~\cite{luo2018fast,hu2021fiery,liang2020pnpnet,djuric2021multixnet,fadadu2022multi,gu2023vip3d,zhang2022beverse}, behavior and planning~\cite{kamenev2022predictionnet,gulino2023waymax,liu2021deep,song2020pip,rhinehart2019precog}, or span from perception to planning~\cite{zeng2019end,sadat2020perceive,casas2021mp3,hu2022st}. Building on the inspiration from \cite{li2022bevformer,zhang2022mutr3d}, Hu \textit{et al.} introduced UniAD~\cite{hu2023planning}, which leverages transformer queries and a shared BEV feature map to facilitate end-to-end learning of perception, prediction, and planning~\cite{chen2023end,jia2023driveadapter,jia2023think}. More recently, there has been a growing interest in achieving end-to-end motion planning using large language models (LLMs)~\cite{xu2023drivegpt4,mao2023gpt,wang2023bevgpt}.

\paragraph{Challenges of Existing Methods} While substantial advancements have been achieved in standard motion prediction benchmarks~\cite{ettinger2021large, caesar2020nuscenes, chang2019argoverse, wilson2023argoverse, chen2023womd}, the deployment of existing behavior models in real-world scenarios remains challenging. 
Many motion prediction models heavily rely on pre-processed, symbolic data from perception models~\cite{qi2021offboard,gao2020vectornet,zhou2018voxelnet,zhou2020end}, and therefore are vulnerable to potential failures. Moreover, the manually-engineered interface greatly restrict the flexibility and scalability of the models in handling long-tail and novel categories of objects. In contrast, end-to-end learning~\cite{chen2023end,hu2023planning,jia2023driveadapter,hagedorn2023rethinking,jia2023think} from raw sensors, while overcoming some limitations, encounters challenges in interpretability and scaling up batch size due to computational constraints.

\section{Multi-modality Scene Tokenization}
\label{sec:method}

We propose a novel method, \modelname{} (\fullmodelname{}), to enrich the information fed to Transformer-based motion prediction models, by efficiently combining existing symbolic representations with scene tokens that encode multi-modality sensor information. In this section, we focus on how we obtain these scene tokens, each represented by a scene element feature enriched with semantic and geometric knowledge extracted from both image and LiDAR data. Figure~\ref{fig:flowchart} shows an overview of MoST.

\subsection{Image Encoding and Point-Pixel Association}
We start by extracting image feature maps for each camera and subsequently associating these features to the corresponding 3D LiDAR points using sensor calibration information. At each time step, we have a set of images $\{\mathbf{I}^k \in \mathbb{R}^{H_k \times W_k \times 3}\}_k$ captured by a total number of $K$ cameras, where $H_k$ and $W_k$ represent the image dimensions. Additionally, we have a LiDAR point cloud $\mathbf{P_{\text{xyz}}} \in \mathbb{R}^{N_{\text{pts}} \times 3}$, with $N_{\text{pts}}$ denoting the number of points. Using a pre-trained 2D image encoder $E^{\text{img}}$, we obtain a feature map of each image, denoted as $\{\mathbf{V}^k \in \mathbb{R}^{H'_k \times W'_k \times D} \}_k$. Subsequently, we leverage camera and LiDAR calibrations to establish a mapping between 3D LiDAR points and their corresponding 2D coordinates on the image feature map of size $H'_k \times W'_k$. This mapping associates each 3D point with the corresponding image feature vector. As a result, we obtain image features for all $N_{\text{pts}}$ 3D points, represented as $\mathbf{F}_{\text{pts}} \in \mathbb{R}^{N_{\text{pts}} \times D}$. Note that for points projecting outside of any image plane, we set their image features as zeros and mark their image features as invalid. 

To harness a wider range of knowledge, we utilize large pre-trained image models trained on a diverse collections of datasets and tasks, capturing a richer understanding of the real world.
We experiment with several image encoder candidates: SAM ViT-H~\cite{kirillov2023segment}, VQ-GAN~\cite{esser2021taming}, CLIP~\cite{radford2021learning} and DINO v2~\cite{oquab2023dinov2}. Different from others, VQ-GAN uses a codebook to build the feature map. To derive $\mathbf{V}^k$ from VQ-GAN, we bottom-crop and partition each input image into multiple 256$\times$256 patches. Subsequently, we extract 256 tokens from each patch and convert them into a 16$\times$16 feature map through querying the codebook. Finally, these partial feature maps are stacked together according to their original spatial locations to produce $\mathbf{V}^k$.

\subsection{Scene Decomposition}
\label{sec:scene_decomposition}

\begin{figure}
\centering
\includegraphics[height=0.6\linewidth]{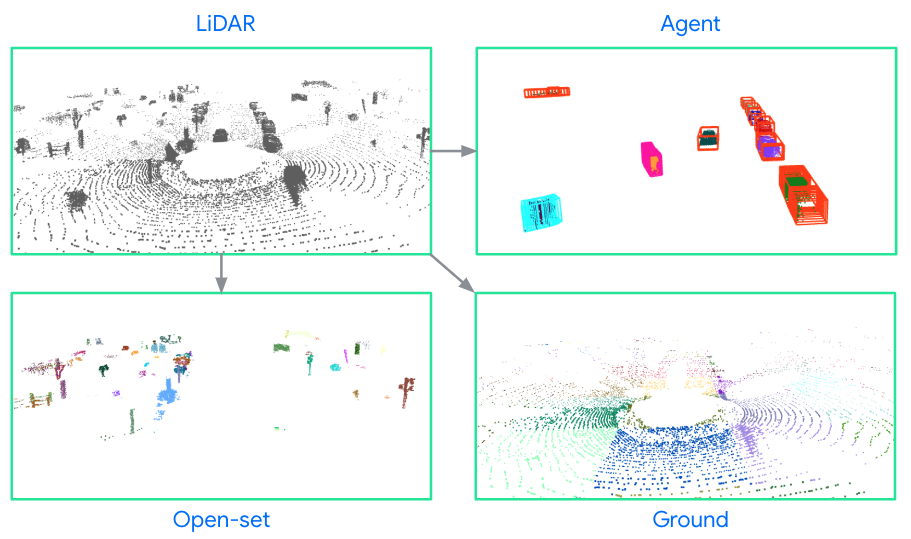}
\cutcaptionup
\cutcaptionup
\cutcaptionup
\caption{Visualization of scene decomposition. We decompose a scene into agent elements, open-set elements and ground elements. We also visualize the perception bounding boxes for agents. }
\label{fig:scene-decomposition}
\end{figure}

Next, our approach groups full scene LiDAR point cloud into three element types:  \textbf{ground}, \textbf{agents}, and \textbf{open-set objects} (see illustration in Figure~\ref{fig:scene-decomposition}). We use the term ``\textbf{scene element}" to denote the union of these three element types. We denote the number of elements of each type to be $N_{\text{elem}}^{\text{gnd}}, N_{\text{elem}}^{\text{agent}}, N_{\text{elem}}^{\text{open-set}}$ respectively, and we define $N_{\text{elem}} =N_{\text{elem}}^{\text{gnd}}+ N_{\text{elem}}^{\text{agent}}+ N_{\text{elem}}^{\text{open-set}}$ as the total number of scene elements.

\begin{itemize}
    \item Ground elements: These are segmented blocks of the ground surface, obtained through either a dedicated ground point segmentation model or a simple RANSAC algorithm. Since the ground occupies a large area, we divide it into disjoint $10m \times 10m$ tiles, following~\cite{liu2022less}.
    \item Agent elements: These correspond to the points within the bounding boxes of agents, detected by established perception pipelines for a pre-defined set of categories.
    \item Open-set object elements: These capture the remaining objects not included in the agent categories. Examples include novel categories of traffic participants and obstacles beyond the training data, long-tail instances that a perception model suppresses due to low confidence. We extract these elements by first removing ground and agent elements from the scene point cloud and then using connected component analysis to group points into instances.
\end{itemize}

\paragraph{Per-point Token ID}
Based on scene decomposition of each LiDAR frame, we can assign a unique token id to each LiDAR point. Points within the same scene element share one token id. With the point-pixel association, we can scatter each scene token ID to a set of camera pixels and/or locations on image feature maps. As a result, we can obtain features from both LiDAR and camera for each scene element.
Based on point-wise token id, we can pool per-point image features into three sets of cluster-wise embedding vectors, \textit{i.e.,} $\mathbf{F}_{\text{img}}^\text{gnd} \in \mathbb{R}^{N_{\text{elm}}^\text{gnd} \times D}$, $\mathbf{F}_{\text{img}}^\text{agent} \in \mathbb{R}^{N_{\text{elm}}^\text{agent} \times D}$, $\mathbf{F}_{\text{img}}^\text{open-set} \in \mathbb{R}^{N_{\text{elm}}^\text{open-set} \times D}$. 

\paragraph{Scene Element Boxes}
We propose to encode each scene element with a combination of image features, coarse-grained geometry features, and fine-grained geometry features. Here we describe how we construct scene element boxes $\mathbf{B}$ to represent coarse-grained geometry. For agent elements, coarse-grained geometry feature are derived from perception pipelines, capturing information of agent positions, sizes, and heading. For open-set object elements, we compute the tightest bounding boxes covering the point cluster, and these bounding boxes are also represented by box centers, box sizes and headings. For ground elements, we have divided the ground into fixed size tiles and simply use the tile center coordinates as position information. These box representations will be further encoded with a MLP and combined with image features and fine-grained features. We will dive into this combination in Sec \ref{sec:scene_element_feature_extraction}.

\subsection{Scene Element Feature Extraction}
\label{sec:scene_element_feature_extraction}

We finally extract scene element features with a neural network module. Multi-frame information are first compressed in an efficient way, then fed into this feature extraction module, which generate a single feature vector for each scene element. The feature extraction module is connected with the downstream Transformer-based motion prediction models, formulating an end-to-end trainable paradigm.

\vspace{-1em}

\subsubsection{Efficient Multi-frame Data Representation}
\label{subsec:efficient_multi_frame_data_representation}

While we've compiled valuable information for each element within a single-frame scene -- LiDAR points, per-point image features, and a bounding box -- collecting this data across multiple frames leads to a large increase in memory usage.
Considering that self-attention layers have quadratic complexity with respect to the number of tokens, naively concatenating tokens across all history frames will also lead to significantly increased memory usage. We propose an efficient data representation to reduce the amount of data sent to the model with the following three ingredients.  

\noindent\textbf{Open-set element tracking} We compress the representation of open-set elements by associating open-set elements across frames using a simple Kalman Filter. For each open-set element, we only store its box information for all $T$ frames with a tensor of shape ($ N_{\text{elem}}^{\text{open-set}}\times T \times 7$) and we apply average pooling across $T$ frames of its image features resulting in an image feature tensor of shape $(N_{\text{elem}}^{\text{open-set}} \times 1 \times D)$.   

\noindent\textbf{Ground-element aggregation} Instead of decomposing the ground into tiles for each frame, we apply decomposition after combining the ground points from all frames.

\noindent\textbf{Cross-frame LiDAR downsampling}
 Directly storing LiDAR points for all frames is computationally prohibitive. Simply downsampling LiDAR points in each frame still suffers from high redundancy over static parts of the scene. Therefore, we employ different downsampling schemes for ground elements (which are always static), and open-set/agent elements (which could be dynamic). For ground elements, we first merge the ground points across all frames then uniformly subsample to a fixed number $N_\text{pts}^\text{gnd}$. For open-set/agent elements, we subsample them to the fixed number $N_\text{pts}^{\text{open-set}}$ and $N_\text{pts}^{\text{agent}}$ respectively. The final LiDAR points $N_{\text{pts}} = N_\text{pts}^{\text{gnd}} + N_\text{pts}^{\text{agent}} + N_\text{pts}^{\text{open-set}}$. We also create a tensor $\mathbf{P}_{\text{ind}} \in \mathbb{R}^{N_{\text{pts}}\times 2}$ that stores the frame id and scene-element id for each point. Note that in this representation, the number of points from each frame is a variable, which is more efficient compared to storing a fixed number of points for all frames with padding.

 \subsubsection{Network Architecture}
 \label{subsec:network_architecture}

With the efficient multi-frame data representation, the inputs to the scene element feature extraction module are summarized as following:
\begin{itemize}
    \item $\mathbf{F}_{\text{pts}} \in \mathbb{R}^{N_{\text{pts}}\times D}$, point-wise image embeddings derived for all the LiDAR points across $T$ frames.
    \item $\mathbf{B} \in \mathbb{R}^{N_{\text{elem}}\times T \times 7}$, bounding boxes of different scene elements across time, where $N_{\text{elem}} = N_{\text{elem}}^{\text{gnd}} + N_{\text{elem}}^{\text{agent}} + N_{\text{elm}}^{\text{open-set}}$. For ground elements, we only encode the tile center, leaving the rest four attributes as zeros. 
    \item $\mathbf{P} =\{ \mathbf{P}_{\text{xyz}}, \mathbf{P}_{\text{ind}} \}$, where $\mathbf{P}_{\text{xyz}} \in \mathbb{R}^{N_{\text{pts}}\times 3}$ collects multi-frame LiDAR points, and $\mathbf{P}_{\text{ind}} \in \mathbb{R}^{N_{\text{pts}}\times 2}$ stores frame id and token id for each point respectively.
\end{itemize}

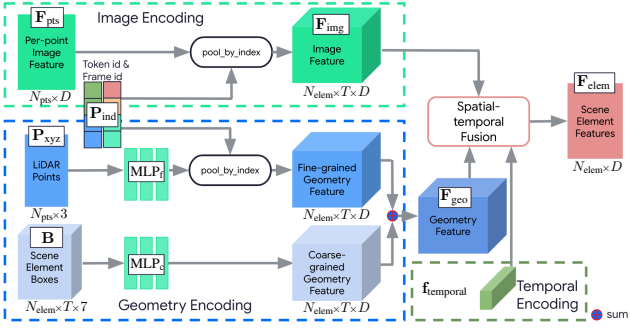
\begin{figure}
\begin{overpic}[width=\linewidth,keepaspectratio]{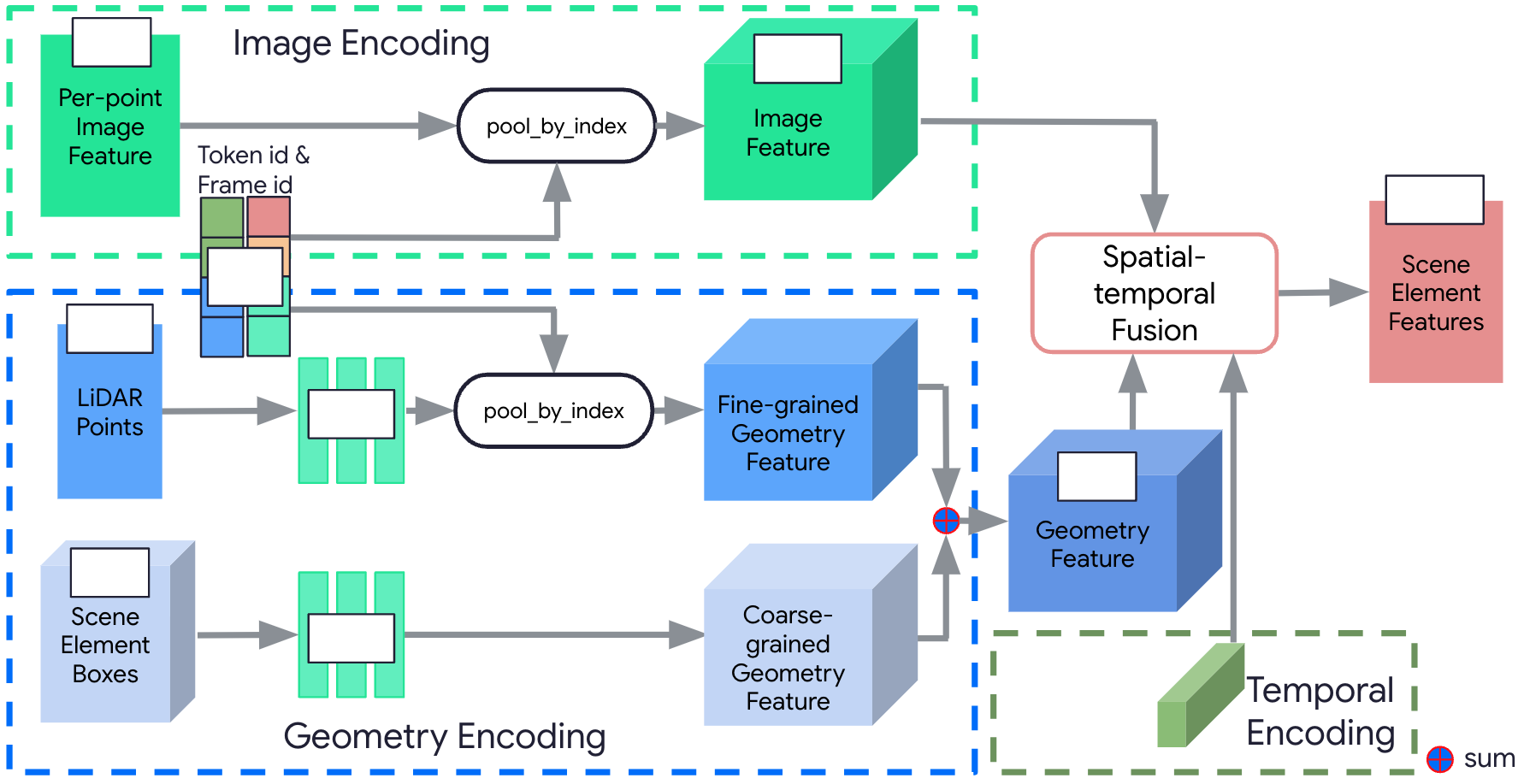}
\put(50,47.2){\scalebox{.6}{$\mathbf{F}_{\text{img}}$}}
\put(5.3,48.3){\scalebox{.6}{$\mathbf{F}_{\text{pts}}$}}
\put(4.8,29.5){\scalebox{.6}{$\mathbf{P}_{\text{xyz}}$}}
\put(6.3,13){\scalebox{.6}{$\mathbf{B}$}}
\put(67,5){\scalebox{.6}{$\mathbf{f}_{\text{temporal}}$}}
\put(69.9,19.8){\scalebox{.6}{$\mathbf{F}_{\text{geo}}$}}
\put(20.5,9){\scalebox{.54}{$\mathbf{\text{MLP}}_{\text{c}}$}}
\put(20.5,23.7){\scalebox{.54}{$\mathbf{\text{MLP}}_{\text{f}}$}}
\put(14,33){\scalebox{.6}{$\mathbf{P}_{\text{ind}}$}}
\put(91.7,38){\scalebox{.6}{{$\mathbf{F}_{\text{elem}}$}}}

\put(4.3,35.7){\scalebox{.53}{$N_{\text{pts}}\plh D$}}
\put(4.5,17){\scalebox{.53}{$N_{\text{pts}}\plh 3$}}
\put(3.7,2.2){\scalebox{.53}{$N_{\text{elem}} \plh T \plh 7$}}
\put(48,36.5){\scalebox{.53}{$N_{\text{elem}}\plh T \plh D$}}
\put(48,16.8){\scalebox{.53}{$N_{\text{elem}}\plh T \plh D$}}
\put(48,2){\scalebox{.53}{$N_{\text{elem}}\plh T \plh D$}}
\put(91,24.5){\scalebox{.53}{$N_{\text{elem}}\plh D$}}

\end{overpic}
\cutcaptionup
\cutcaptionup
\caption{Scene element feature extraction. Scene-element feature is derived from a spatial-temporal module that fusing together image feature, geometry feature and temporal embedding. Image feature contains pooled feature from large pre-trained image encoder, and characterize the appearance and semantic attribute of the scene element. Geometry feature, on the other hand, characterizes the spatial location as well as the detailed geometry. Temporal information is injected through a learned temporal embedding. }
\label{fig:tokenizer}
\end{figure}

For each tracked element across $T$ time steps, our network (as shown in Figure~\ref{fig:tokenizer}) will process the previously listed multi-modality information into one embedding for each scene element, denoted as $\mathbf{F}_{\text{elem}}$. 

As shown in the top branch of Figure~\ref{fig:tokenizer}, the network leverages $\mathbf{P}_{\text{ind}}$ to group point-wise image embeddings $\mathbf{F}_{\text{pts}}$ according to the token id and frame id, which results in the image feature tensor for all scene elements across frames, $\mathbf{F}_{\text{img}}$. In the bottom branch of Figure~\ref{fig:tokenizer}, we aim to derive geometry information $\mathbf{F}_{\text{geo}}$ by encoding two pieces of information, \textit{i.e.,} fine-grained geometry information from point clouds $\mathbf{P}_{\text{xyz}}$ and coarse-grained shape information from 3D boxes $\mathbf{B}$. The fine-grained geometry is encoded by first mapping point xyz coordinates into a higher dimensional space and grouping high-dimensional features according to the token id and frame id. The coarse-grained shape encoding is derived by projecting box attributes to the same high dimensional space. 
Formally, $\mathbf{f}_{\text{geo}}$ is defined as 

\begin{align}
\begin{split}
    \mathbf{f}_{\text{geo}}^{i} &= \text{pool\_by\_index}
    (\text{MLP}_f(\mathbf{P}_{\text{xyz}}), \mathbf{P}_{\text{ind}})[i, :, :] \\
     &+ \text{MLP}_c(\mathbf{B})[i, :, :]
\end{split}
\end{align} 
where $i$ is the token id, function \textit{$\text{pool\_by\_index}$} pools point-wise features based on token id and frame id.

\noindent\textbf{Spatial-temporal Fusion}  Our spatial-temporal fusion module (Figure~\ref{fig:tokenizer} right) takes as input the image feature $\mathbf{F}_{\text{img}}$, the geometry feature $\mathbf{F}_{\text{geo}}$, and a trainable temporal embedding $\mathbf{f}_{\text{temporal}}\in \mathbb{R}^{T\times D}$ that corresponds to $T$ frames. It produces a temporally aggregated feature $\mathbf{F}_{\text{elem}}$ for all scene elements. Under the hood, the spatial-temporal fusion module adds up the two input tensors, and then
conducts axial attention across the temporal and element axes of the tensor, which is followed by the final average pooling across the temporal axis, as listed below:
\begin{align}
\begin{split}
\nonumber \mathbf{F}_{\text{elem}} &\gets \mathbf{F}_{\text{img}} + \mathbf{F}_{\text{geo}} + \mathbf{f}_\text{temporal}\in \mathbb{R}^{N_{\text{elem}} \times T \times D} \\ 
\mathbf{F}_{\text{elem}} &\gets \text{AttnAlongAxis}(\mathbf{F}_{\text{elem}}, \text{axis}=\text{time}) \\ 
\mathbf{F}_{\text{elem}} &\gets \text{AttnAlongAxis}(\mathbf{F}_{\text{elem}}, \text{axis}=\text{scene element}) \\
\mathbf{F}_{\text{elem}} &\gets \text{mean}(\mathbf{F}_{\text{elem}}, \text{axis}=\text{time})
\end{split}
\end{align}
The final $\mathbf{F}_{\text{elem}}$ uses a single vector to describe each scene element. This tensor can be fed as the additional inputs to scene encoding module of transformer-based motion prediction models, such as recently published ~\cite{nayakanti2023wayformer,shi2022motion}.

\section{Experiments}
\label{sec:Experiments}
\begin{table}
\setlength\tabcolsep{7pt}
    \centering
\begin{tabular}{r|cccc}
\cline{1-3}
\multicolumn{1}{l|}{} & \multicolumn{1}{c|}{ViT-VQGAN~\cite{yu2021vector}}         & \multicolumn{1}{c}{SAM ViT-H~\cite{kirillov2023segment}}        &  &  \\ \cline{1-3}
Sensors        & \multicolumn{2}{c}{\begin{tabular}[c]{@{}c@{}}8 cameras (front, front left \\ front right, side left, side right,\\ rear left, rear right, rear)  and LiDAR\end{tabular}} &  &  \\ \cline{1-3}
Temporal        & \multicolumn{2}{c}{1.0 s, 11 Frames}                                                                                                                                                  &  &  \\ \cline{1-3}
\begin{tabular}[c]{@{}r@{}}Pre-trained\\ Dataset\end{tabular}              & \multicolumn{1}{c|}{WebLi~\cite{chen2022pali}}            & SA-1B~\cite{kirillov2023segment}                                &  &  \\ \cline{1-3}
Format                & \multicolumn{1}{c|}{Token \& Embedding}                                                      & Embedding                                                                             \\ \cline{1-3}\cline{1-3}
\end{tabular}
\caption{Details of the WOMD camera embeddings.}
\label{tbl:dataset_release}
\end{table}

\begin{table*}[t]
    \setlength\tabcolsep{4pt}
    \centering
    \begin{tabular}{llc|c|ccc|cc}
        \toprule
        Method & Reference & Sensor & \# Decoders & \multicolumn{1}{c}{minADE$\downarrow$} & minFDE$\downarrow$ &Miss Rate$\downarrow$& \multicolumn{1}{c}{mAP$\uparrow$} & soft-mAP$\uparrow$ \\
        \midrule
        MotionCNN~\cite{Motioncnn2021} & CVPRW 2021 & - & - & 0.7383 & 1.4957 & 0.2072 & 0.2123 & - \\
        MultiPath++~\cite{ngiam2022scenetransformer} & ICRA 2022 & - & -  & 0.978 & 2.305 & 0.440 & - & - \\
        SceneTransformer~\cite{ngiam2022scenetransformer} & ICLR 2022 & - & -  & 0.9700 & 2.0700 & 0.1867 & 0.2433 & - \\       
        MTR~\cite{shi2022motion} & NeurIPS 2022 & -  & - & 0.6046  & 1.2251 & 0.1366 & 0.4164 &- \\
        Wayformer~\cite{nayakanti2023wayformer} & ICRA 2023 & - & 3  & 0.5512  & 1.1602 & 0.1208 & 0.4099 & 0.4247 \\ 
        MotionLM*~\cite{seff2023motionlm} &  ICCV 2023 & - & 1 & 0.5702  & 1.1653 & 0.1327 &0.3902 &0.4063 \\
        \midrule
        Wayformer & Reproduced & - & 1  &  0.5830  & 1.2314 & 0.1347 & 0.3995 & 0.4110 \\  
        \modelname{}-SAM\_H-6 & Ours & C+L & 1  & \textbf{0.5228} & \textbf{1.0764} & 0.1303 & 0.4040 & 0.4207 \\
        \modelname{}-SAM\_H-64 & Ours & C+L & 1  & 0.5487 & 1.1355 & 0.1238 & \textbf{0.4230} & \underline{0.4380} \\
        \midrule
        Wayformer & Reproduced  & - & 3  & 0.5494  & 1.1386 & \underline{0.1190} & 0.4052 & 0.4239 \\
        MoST-VQGAN-64 & Ours & C & 3  & \underline{0.5391} & \underline{1.1099} & \textbf{0.1172} & \underline{0.4201} & \textbf{0.4396} \\   
        \bottomrule
        \bottomrule %
    \end{tabular}
    \cutcaptionup
    
    \caption{Performance comparison on WOMD validation set. \modelname{} leads to significant performance gain to the Wayformer baselines and achieves state-of-the-art results in all compared metrics. \modelname{}-SAM\_H-$\{6,64\}$: our method using SAM ViT-H feature and predicting based on 6 or 64 queries. \modelname{}-VQGAN-64: our method using VQGAN feature with 64 queries. Bold font highlights the best result in each metric and underline denotes the second best. 
    For methods with multiple decoders, results are based on ensembling of predictions. MotionLM* is based on contacting authors for their 1 decoder results, which was not reported in the original publication.}
    \label{tbl:main}
    \vspace{-1em}
\end{table*}

\subsection{The Release of WOMD Camera Embeddings}
To advance research in sensor-based motion prediction, we have augmented Waymo Open Motion Dataset (WOMD)~\cite{Ettinger_2021_ICCV} with camera embeddings.
WOMD contains the standard perception output,
e.g. tracks of bounding boxes, road graph, traffic signals, and now it also includes synchronized LiDAR points~\cite{Ettinger_2021_ICCV} and camera embeddings.
Given one scenario, a motion prediction model is required to reason about 1 second history data and generate predictions for the future 8 seconds at 5Hz. 
Our LiDAR can reach up to 75 meters along the radius and the cameras provide a multi-view imagery for the environment. WOMD characterize each perception-detected objects using a 3D bounding box (3D center point, heading, length, width, and height), and the object’s velocity vector. 
The road graph is provided as a set of polylines and polygons with semantic types. WOMD is divided into training, validation and testing subsets according to the ratio 70\%, 15\%, 15\%. 
In this paper, we report results over the validation set. We'll reserve the test set for future community benchmarking.

Due to the data storage issue and risk of leakage of sensitive information (e.g., human faces, car plate numbers, etc.), we will not release the raw camera images. Instead, the released multi-modality dataset will be in two formats:
\begin{itemize}
    \item \textbf{ViT-VQGAN Tokens and Embeddings}: We apply a pre-trained ViT-VQGAN~\cite{yu2021vector} to extract tokens and embeddings for each camera image. The number of tokens per camera is 512, where each token corresponds to a 32 dimensional embedding in the quantized codebook.
    \item \textbf{SAM ViT-H Embeddings}: We apply a pre-trained SAM ViT-H~\cite{kirillov2023segment} model to extract dense embeddings for each camera image. We release the per-scene-element embedding vectors, each being 256 dimensional.
\end{itemize}
In the released dataset, we have 1 LiDAR and 8 cameras (front, front-left, front-right, side-left, side-right, rear-left, rear-right, rear). 
Please see details in Table~\ref{tbl:dataset_release}.

\paragraph{Task and Metrics}
Based on the augmented WOMD, we investigate the standard marginal motion prediction task, where a model is required to generate 6 mostly likely future trajectories for each of the agents independently of other agents futures.
We report results for various methods under commonly adopted metrics, namely minADE, minFDE, miss rate, mAP and soft-mAP~\cite{Ettinger_2021_ICCV}.
For fair comparison, we only compare results based on single model prediction.

\subsection{Experimental Results}
Our \modelname{} is a general paradigm applicable to most transformer-based motion prediction architectures. Without losing generality, we adopt a state-of-the-art architecture, Wayformer~\cite{nayakanti2023wayformer}, as our motion prediction backbone and we augment it with our new design by fusing multi-modality tokens. In the following sections, we use Wayformer as the baseline and show the performance improvement by \modelname{}. Please refer to appendix for implementation details.

\subsubsection{Baseline Comparison}
\label{subsubsec:baseline_comparison}
In Table~\ref{tbl:main}, we evaluate the proposed approach and compare it with recently published models, \textit{i.e.,} MTR~\cite{shi2022motion}, Wayformer~\cite{nayakanti2023wayformer}, MultiPath++~\cite{multipath_plusplus2022}, MotionCNN~\cite{Motioncnn2021}, 
MotionLM~\cite{seff2023motionlm}, SceneTransformer~\cite{ngiam2022scenetransformer}. 
Specifically, we study our approach in two settings, 1) using LiDAR + camera tokens with single decoder and 2) using camera tokens with 3 decoders. During inference, we fit a Gaussian Mixture Model by merging predictions from the decoder(s) and draw 2048 samples, which are finally aggregated into 6 trajectories through K-means~\cite{nayakanti2023wayformer}. In both settings, the introduction of sensory tokens leads to a clear performance gain over the corresponding Wayformer baselines based on our re-implementation. Moreover, our approach achieves state-of-the-art performance across various metrics.

\begin{figure}[ht]
  \centering
  \includegraphics[width=\linewidth]{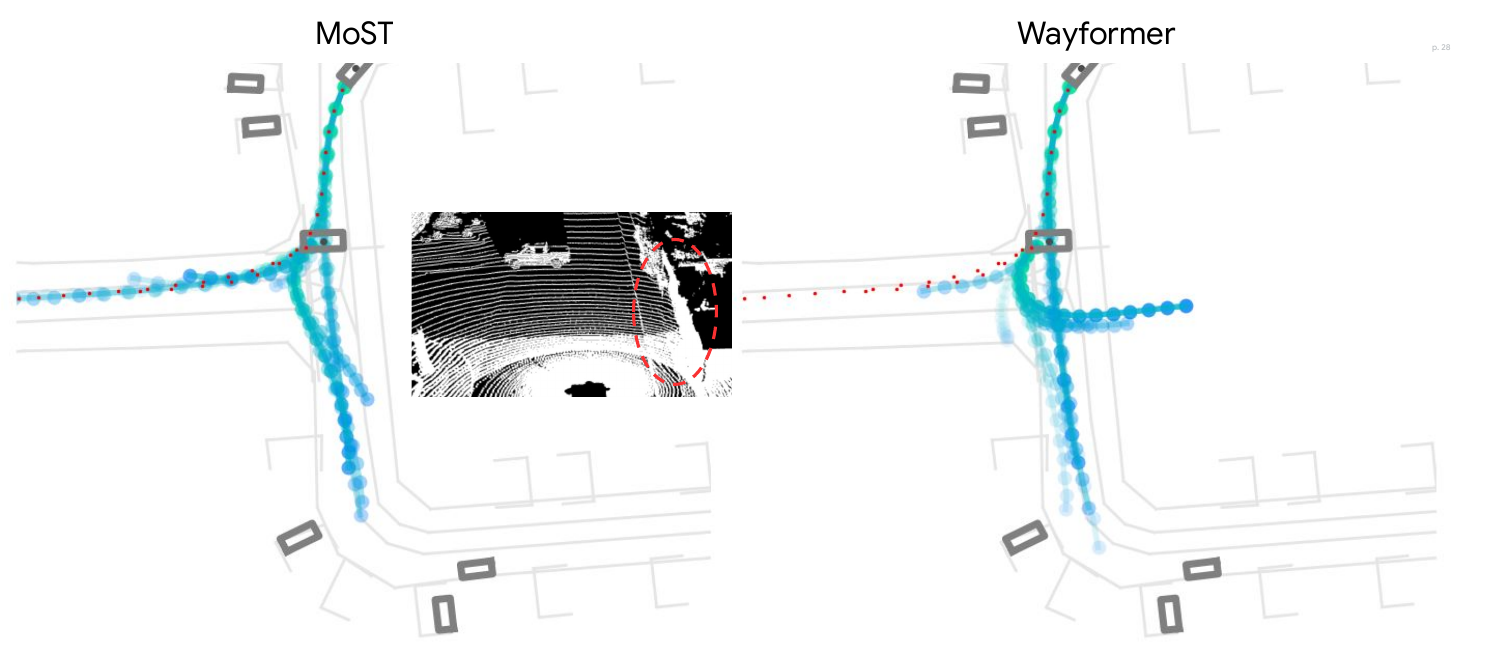}
  \includegraphics[width=\linewidth]{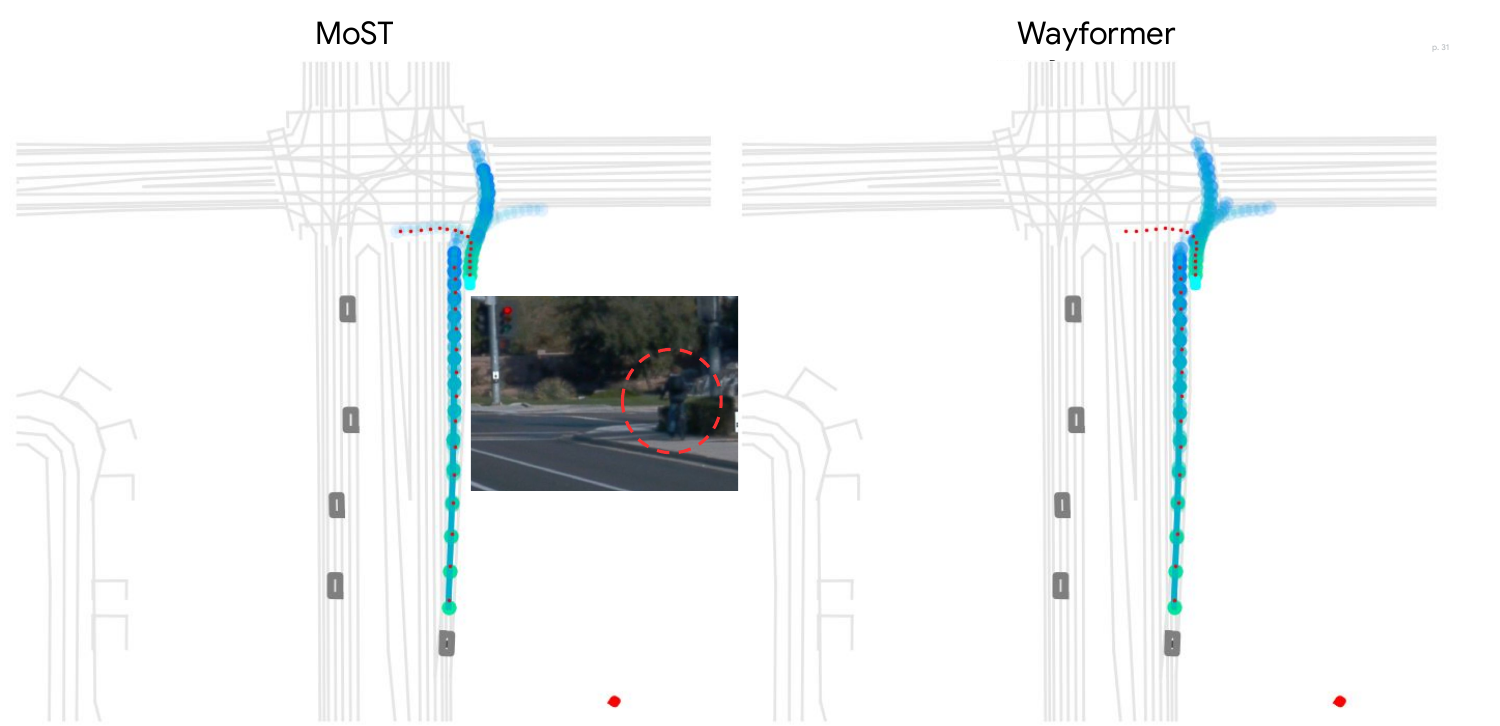}
  \caption{Qualitative comparison. The agent boxes are colored by their types: gray for vehicle, red for pedestrian, and cyan for cyclist. The predicted trajectories are ordered temporally from green to blue. For each modeled agent, the models predict 6 trajectory candidates, whose confidence scores are illustrated by transparency: the more confident, the more visible. Ground truth trajectory is shown as red dots. In the upper example, \modelname{} rules out the possibility that a vehicle runs onto a wall after U-turn; in the lower example, \modelname{} correctly predicts that a cyclist could suddenly cross the street.}
  \label{fig:qual_comparison}
\end{figure}

Figure~\ref{fig:qual_comparison} illustrates two comparisons between our \modelname{} and the baseline model. The upper example shows that with tokenized sensor information, our \modelname{} rules out the possibility that a vehicle runs onto walls after a U-turn. The lower example makes a prediction that a cyclist may cross the street which is safety critical for the autonomous vehicle to take precaution regarding this behavior.

\begin{table}
    \setlength\tabcolsep{7pt}
    \centering
    \begin{tabular}{l|ccc}
        \toprule
        Image Encoder & \multicolumn{1}{c}{minADE$\downarrow$} &  \multicolumn{1}{c}{mAP$\uparrow$} & soft-mAP$\uparrow$ \\
        \midrule
        DINO-v2~\cite{oquab2023dinov2}& 0.5597  & 0.4154  & 0.4285 \\
        CLIP~\cite{radford2021learning}& 0.5590  & 0.4138 & 0.4272  \\
        VQ-GAN~\cite{esser2021taming}& 0.5670 & 0.4058 & 0.4192 \\
        SAM ViT-H~\cite{kirillov2023segment}& \textbf{0.5483} & \textbf{0.4162}  & \textbf{0.4321} \\
        \bottomrule
    \end{tabular}
    
    \caption{Ablation study of different image features. All these image features improves the motion prediction performance, while we observe SAM ViT-H~\cite{kirillov2023segment} leads to the most improvement. We uses single-frame multi-modal feature for these study.}
    \label{tbl:ablation:feature}
\end{table}

\subsubsection{Ablation Study}
We find that applying MoST to the current frame can also lead to significant improvement over the baseline. For efficient experimentation, we perform ablation study by employing single frame MoST and SAM ViT-H feature.

\noindent\textbf{Effects of Different Pre-trained Image Encoders} To investigate different choices of the image encoder for our model, we have conducted experiments comparing the performance of using image feature encoders from various pre-trained models: SAM ViT-H~\cite{kirillov2023segment}, CLIP~\cite{radford2021learning}, DINO v2~\cite{oquab2023dinov2}, and VQ-GAN~\cite{esser2021taming}. As show in Table~\ref{tbl:ablation:feature}, the SAM ViT-H encoder yields the highest performance across all behavior prediction metrics. We hypothesize that this performance advantage likely stems from SAM's strong capability to extract comprehensive and spatially faithful feature maps, as it is trained with a large and diverse dataset for the dense understanding task of image segmentation. Other large pre-trained image models also demonstrate notable capability, outperforming the Wayformer baseline on mAP and soft mAP, albeit inferior to SAM.

\begin{table}
    \setlength\tabcolsep{2pt}
    \centering
    \begin{tabular}{cccc|ccc}
        \toprule
         Open-set & Agent & Ground & M-frame&\multicolumn{1}{c}{minADE$\downarrow$} &  soft-mAP$\uparrow$ \\
        \midrule
        \xmark& \cmark & \cmark& \xmark & 0.5654  & 0.4112\\        
        \cmark& \xmark& \cmark & \xmark& 0.5520 & 0.4273\\
        \cmark & \cmark& \xmark& \xmark& 0.5514 &0.4241\\
        \cmark & \cmark& \cmark& \xmark&\textbf{0.5483}  &0.4321\\
        \cmark & \cmark& \cmark& \cmark& 0.5487   &\textbf{0.4380}\\
        \bottomrule
    \end{tabular}
    \caption{Ablation study on how different scene element affects the performance. The first four rows shows that all types of scene element brings benefits to the model. The last row shows that aggregating scene element across frame considerably improves the soft-mAP, though leads to slight regression of minADE. }
    \label{tbl:ablation:element}
\end{table}

\begin{table}
    \setlength\tabcolsep{1.5pt}
    \centering
    \begin{tabular}{ccc|ccc}
        \toprule
         Perception & Camera & LiDAR &\multicolumn{1}{c}{minADE$\downarrow$} &  mAP$\uparrow$ &soft-mAP$\uparrow$ \\
        \midrule
        \cmark& \xmark& \xmark & 0.5830 & 0.3995 & 0.4110 \\
        \cmark& \cmark& \xmark & \textbf{0.5483}  &0.4118 &  0.4265\\
        \cmark& \xmark& \cmark & 0.5486 & 0.4040 & 0.4212  \\
        \cmark& \cmark& \cmark & \textbf{0.5483} & \textbf{0.4162}& \textbf{0.4321} \\
        \bottomrule
    \end{tabular}
    
    \caption{Ablation study on different input modality. The first row corresponds to the setting of Wayformer~\cite{nayakanti2023wayformer}. The second row adds camera image feature. The last row further adds LiDAR. We can see using both Camera and LiDAR yields to the best results.}
    \label{tbl:ablation:input}
\end{table}

\begin{table}
    \setlength\tabcolsep{1.5pt}
    \centering
    \begin{tabular}{l|ccc}
        \toprule
         Sensory Token  &\multicolumn{1}{c}{minADE$\downarrow$} &  mAP$\uparrow$ &soft-mAP$\uparrow$ \\
        \midrule
        None (Wayformer) & 0.5830& 0.3995& 0.4110\\
        Image-grid Token (Ours) & 0.5495& 0.4109&0.4261\\
        Scene Cluster Token (Ours) & \textbf{0.5483} & \textbf{0.4162}& \textbf{0.4321} \\
        \bottomrule
    \end{tabular}
    \caption{Comparing variant of scene tokenization strategy with single-frame sensor data. Both token strategy leads to improvement the vanilla Wayformer~\cite{nayakanti2023wayformer}, which does not use sensor data. }
    \label{tbl:ablation:token}
\end{table}

\noindent\textbf{Ablation on Input Modality} To understand how different input modalities affect the final model performance, we conduct ablation experiments and summarize results in Table~\ref{tbl:ablation:input} where we remove image feature or remove LiDAR feature of our single-frame model. We can see image feature and LiDAR feature are both beneficial, and combining both modality leads to the biggest improvement.

\noindent\textbf{Ablation on Scene Element} To gain deeper insights into the contribution of each type of scene element, we conduct ablation studies in Table~\ref{tbl:ablation:element} by removing specific element types and evaluate the impact on behavior prediction metrics. Combining all types of scene elements leads to the best soft-mAP metric. Note that only associating image features to agents give the smallest improvement. We hypothesis the reasons to be two-fold: 1) in most cases, the agent box is sufficient to characterize the motion of the object; 2) there are only a handful of agents in the scene and very few image features are included in the model.  

\noindent\textbf{Alternative Scene Tokenizer} We use SAM ViT-H in this experiment. We design another baseline tokenizer, denoted as Image-grid token, which tokenizes each image feature as $16 \times 16 = 256$ image embeddings by subsampling $4$X along column and row axes. The feature from all camera images are flattened and concatenated to form scene tokens. In Table~\ref{tbl:ablation:token} we can see the Image-grid tokenizer also leads to improvement compared to Wayformer baseline, though inferior to our cluster-based sparse tokenizer which utilizes point cloud to derive accurate depth information and leverages the intrinsic scene sparsity to get compact tokens.

\subsubsection{Evaluation on Challenging Scenarios} 
While the improvement shown above demonstrates overall improvement across all driving scenarios, we are also interested in investigating the performance gain in the most challenging cases.
Here we present how our model performs in challenging scenarios, specifically on (a) a mined set of hard scenarios, (b) situations where perception failures happen, and (c) situations where roadgraph is inaccurate.

\noindent\textbf{Mined Hard Scenarios} 
To assess the effectiveness of our method in complex situations, we have curated a set of hard scenarios. We conduct a per-scenario evaluation throughout the entire validation set, identifying the 1000 scenarios with the lowest minADE across vehicle, pedestrian, and cyclist categories for the baseline and \modelname{}-SAM\_H-64, respectively. In this way, we ensure the mining is symmetric and fair for both methods. Then we combine these 6000 scenarios, resulting in 4024 unique scenarios, forming our curated challenging evaluation dataset. 
As shown in Table~\ref{tbl:hard_scenarios}, \modelname{} demonstrates more pronounced relative improvement in mAP and soft-mAP, \textit{i.e.,} $13.1\%$ and $12.4\%$ respectively, compared to the baseline in these hardest scenarios, confirming its effectiveness of enhanced robustness and resilience in complex situations. We also find that improving minADE in these hard scenarios is still a challenge.

\begin{table}[t]
    \begin{tabular}{l|ccc}
    \hline
    Method    & minADE$\downarrow$ & mAP$\uparrow$    & soft-mAP$\uparrow$ \\ \hline
    Wayformer~\cite{nayakanti2023wayformer} & 0.9002 & 0.2312 & 0.2382   \\ \hline
    \modelname-SAM\_H-64      & \textbf{0.8720} & \textbf{0.2615} & \textbf{0.2677}   \\ \hline
    \end{tabular}
    \cutcaptionup
    \caption{Evaluation on hard scenarios. We curate a set of hard scenarios based on the performance of \modelname{} and Wayformer on them. \modelname{} consistently shows improved performance.}
    \label{tbl:hard_scenarios}
\end{table}

\noindent\textbf{Perception Failure} Most motion prediction algorithms~\cite{nayakanti2023wayformer,seff2023motionlm} assume accurate perception object boxes as inputs. It is critical to understand how such a system will perform when this assumption breaks due to various reasons, such as long-tail and novel categories of object beyond training supervision, occlusion, long-range, etc. Thus, we propose to additionally evaluate our method against the baseline method in the case of perception failure. Concretely, we simulate perception failure of not detecting certain object boxes by randomly removing agents according to a fixed ratio of agents in the scene. The boxes dropped out are consistent for \modelname-SAM\_H-64 and the baseline. As shown in Table~\ref{tbl:perception-roadgraph-failure}, \modelname{} shows robustness against perception failures: even when failure rate raises to 50\%, our model still performs on par with the Wayformer baseline (soft mAP 0.4121).

\noindent\textbf{Roadgraph Failure} Motion prediction models often exhibit a strong reliance on roadgraphs, leading to potential vulnerabilities in situations where the roadgraph is incomplete or inaccurate. Our proposed model, \modelname{}, tackles this issue by incorporating multi-modality scene tokens as additional inputs, thereby enhancing its robustness against roadgraph failures. We demonstrate this advantage by simulating various levels of roadgraph errors, similar to the aforementioned perception failure simulation. Specifically, we evaluate \modelname-SAM\_H-64 under scenarios with 10\%, 30\%, and 50\% missing roadgraph segments in the validation set. Notably, as showcased in Table~\ref{tbl:perception-roadgraph-failure}, even with a 30\% of the roadgraph missing, \modelname{} performs on par with baseline models that assume perfect roadgraph information.

\begin{table}[t]
    \centering
    \begin{tabular}{l|c|cc}
    \hline
    Failure type                 & Failure rate & \multicolumn{1}{l}{minADE$\downarrow$}     & \multicolumn{1}{l}{soft-mAP$\uparrow$}   \\ \hline
           None                & 0\%          & 0.5515                         & 0.4396  \\ \hline
                           & 10\%         & 0.5560                         &  0.4302                         \\
    \multirow{-1}{*}{Perception}  & 30\%         & 0.5625                         & 0.4235                         \\
                           & 50\%         & 0.5712                         & 0.4164                         \\ \hline
    & 10\% & 0.5647 & 0.4217 \\
    \multirow{-1}{*}{Roadgraph}  & 30\% & 0.6020 & 0.4010 \\
    & 50\% & 0.6707 & 0.3499 \\ \hline
    \end{tabular}
    \caption{Evaluation on simulated perception and roadgraph failure. We vary the ratio of miss detected agent boxes and miss detected roadgraph segments in scenes as 10\%, 30\% and 50\%, respectively. With multi-modal features, \modelname{} performs on par with baselines even with 50\% perception or 30\% roadgraph failure.}
    \label{tbl:perception-roadgraph-failure}
\end{table}

\section{Conclusions}
 To promote sensor-based motion prediction research, we have enhanced WOMD with camera embeddings, making it a large-scale multi-modal dataset for benchmarking. To efficiently integrate multi-modal sensor signals into motion prediction, we propose a method that represents the multi-frame scenes as a set of scene elements and leverages large pre-trained image encoders and 3D point cloud networks to encode rich semantic and geometric information for each element.  We demonstrate that our approach leads to significant improvements in motion prediction task.
{
    \small
    \bibliographystyle{ieeenat_fullname}
    \bibliography{main}

\begin{thebibliography}{75}
\providecommand{\natexlab}[1]{#1}
\providecommand{\url}[1]{\texttt{#1}}
\expandafter\ifx\csname urlstyle\endcsname\relax
  \providecommand{\doi}[1]{doi: #1}\else
  \providecommand{\doi}{doi: \begingroup \urlstyle{rm}\Url}\fi

\bibitem[Biktairov et~al.(2020)Biktairov, Stebelev, Rudenko, Shliazhko, and Yangel]{biktairov2020prank}
Yuriy Biktairov, Maxim Stebelev, Irina Rudenko, Oleh Shliazhko, and Boris Yangel.
\newblock Prank: motion prediction based on ranking.
\newblock \emph{Advances in neural information processing systems}, 33:\penalty0 2553--2563, 2020.

\bibitem[Buhet et~al.(2020)Buhet, Wirbel, Bursuc, and Perrotton]{buhet2020plop}
Thibault Buhet, Emilie Wirbel, Andrei Bursuc, and Xavier Perrotton.
\newblock Plop: Probabilistic polynomial objects trajectory planning for autonomous driving.
\newblock \emph{arXiv preprint arXiv:2003.08744}, 2020.

\bibitem[Caesar et~al.(2020)Caesar, Bankiti, Lang, Vora, Liong, Xu, Krishnan, Pan, Baldan, and Beijbom]{caesar2020nuscenes}
Holger Caesar, Varun Bankiti, Alex~H Lang, Sourabh Vora, Venice~Erin Liong, Qiang Xu, Anush Krishnan, Yu Pan, Giancarlo Baldan, and Oscar Beijbom.
\newblock nuscenes: A multimodal dataset for autonomous driving.
\newblock In \emph{CVPR}, 2020.

\bibitem[Casas et~al.(2018)Casas, Luo, and Urtasun]{casas2018intentnet}
Sergio Casas, Wenjie Luo, and Raquel Urtasun.
\newblock Intentnet: Learning to predict intention from raw sensor data.
\newblock In \emph{CoRL}, 2018.

\bibitem[Casas et~al.(2020)Casas, Gulino, Liao, and Urtasun]{casas2020spagnn}
Sergio Casas, Cole Gulino, Renjie Liao, and Raquel Urtasun.
\newblock Spagnn: Spatially-aware graph neural networks for relational behavior forecasting from sensor data.
\newblock In \emph{2020 IEEE International Conference on Robotics and Automation (ICRA)}, pages 9491--9497. IEEE, 2020.

\bibitem[Casas et~al.(2021)Casas, Sadat, and Urtasun]{casas2021mp3}
Sergio Casas, Abbas Sadat, and Raquel Urtasun.
\newblock Mp3: A unified model to map, perceive, predict and plan.
\newblock In \emph{Proceedings of the IEEE/CVF Conference on Computer Vision and Pattern Recognition}, pages 14403--14412, 2021.

\bibitem[Chai et~al.(2019{\natexlab{a}})Chai, Sapp, Bansal, and Anguelov]{chai2019multipath}
Yuning Chai, Benjamin Sapp, Mayank Bansal, and Dragomir Anguelov.
\newblock Multipath: Multiple probabilistic anchor trajectory hypotheses for behavior prediction.
\newblock In \emph{CoRL}, 2019{\natexlab{a}}.

\bibitem[Chai et~al.(2019{\natexlab{b}})Chai, Sapp, Bansal, and Anguelov]{multipath_2019}
Yuning Chai, Benjamin Sapp, Mayank Bansal, and Dragomir Anguelov.
\newblock Multipath: Multiple probabilistic anchor trajectory hypotheses for behavior prediction.
\newblock In \emph{CoRL}, 2019{\natexlab{b}}.

\bibitem[Chang et~al.(2019)Chang, Lambert, Sangkloy, Singh, Bak, Hartnett, Wang, Carr, Lucey, Ramanan, et~al.]{chang2019argoverse}
Ming-Fang Chang, John Lambert, Patsorn Sangkloy, Jagjeet Singh, Slawomir Bak, Andrew Hartnett, De Wang, Peter Carr, Simon Lucey, Deva Ramanan, et~al.
\newblock Argoverse: 3d tracking and forecasting with rich maps.
\newblock In \emph{CVPR}, 2019.

\bibitem[Chen et~al.(2023{\natexlab{a}})Chen, Ge, Qiu, Al-Rfou, Qi, Zhou, Yang, Ettinger, Sun, Leng, et~al.]{chen2023womd}
Kan Chen, Runzhou Ge, Hang Qiu, Rami Al-Rfou, Charles~R Qi, Xuanyu Zhou, Zoey Yang, Scott Ettinger, Pei Sun, Zhaoqi Leng, et~al.
\newblock Womd-lidar: Raw sensor dataset benchmark for motion forecasting.
\newblock \emph{arXiv preprint arXiv:2304.03834}, 2023{\natexlab{a}}.

\bibitem[Chen et~al.(2023{\natexlab{b}})Chen, Wu, Chitta, Jaeger, Geiger, and Li]{chen2023end}
Li Chen, Penghao Wu, Kashyap Chitta, Bernhard Jaeger, Andreas Geiger, and Hongyang Li.
\newblock End-to-end autonomous driving: Challenges and frontiers.
\newblock \emph{arXiv preprint arXiv:2306.16927}, 2023{\natexlab{b}}.

\bibitem[Chen et~al.(2022)Chen, Wang, Changpinyo, Piergiovanni, Padlewski, Salz, Goodman, Grycner, Mustafa, Beyer, et~al.]{chen2022pali}
Xi Chen, Xiao Wang, Soravit Changpinyo, AJ Piergiovanni, Piotr Padlewski, Daniel Salz, Sebastian Goodman, Adam Grycner, Basil Mustafa, Lucas Beyer, et~al.
\newblock Pali: A jointly-scaled multilingual language-image model.
\newblock \emph{arXiv preprint arXiv:2209.06794}, 2022.

\bibitem[Cui et~al.(2019)Cui, Radosavljevic, Chou, Lin, Nguyen, Huang, Schneider, and Djuric]{cui2019multimodal}
Henggang Cui, Vladan Radosavljevic, Fang-Chieh Chou, Tsung-Han Lin, Thi Nguyen, Tzu-Kuo Huang, Jeff Schneider, and Nemanja Djuric.
\newblock Multimodal trajectory predictions for autonomous driving using deep convolutional networks.
\newblock In \emph{ICRA}, 2019.

\bibitem[Djuric et~al.(2021)Djuric, Cui, Su, Wu, Wang, Chou, San~Martin, Feng, Hu, Xu, et~al.]{djuric2021multixnet}
Nemanja Djuric, Henggang Cui, Zhaoen Su, Shangxuan Wu, Huahua Wang, Fang-Chieh Chou, Luisa San~Martin, Song Feng, Rui Hu, Yang Xu, et~al.
\newblock Multixnet: Multiclass multistage multimodal motion prediction.
\newblock In \emph{2021 IEEE Intelligent Vehicles Symposium (IV)}, pages 435--442. IEEE, 2021.

\bibitem[Esser et~al.(2021)Esser, Rombach, and Ommer]{esser2021taming}
Patrick Esser, Robin Rombach, and Bjorn Ommer.
\newblock Taming transformers for high-resolution image synthesis.
\newblock In \emph{Proceedings of the IEEE/CVF conference on computer vision and pattern recognition}, pages 12873--12883, 2021.

\bibitem[Ettinger et~al.(2021{\natexlab{a}})Ettinger, Cheng, Caine, Liu, Zhao, Pradhan, Chai, Sapp, Qi, Zhou, Yang, Chouard, Sun, Ngiam, Vasudevan, McCauley, Shlens, and Anguelov]{Ettinger_2021_ICCV}
Scott Ettinger, Shuyang Cheng, Benjamin Caine, Chenxi Liu, Hang Zhao, Sabeek Pradhan, Yuning Chai, Ben Sapp, Charles~R. Qi, Yin Zhou, Zoey Yang, Aur\'elien Chouard, Pei Sun, Jiquan Ngiam, Vijay Vasudevan, Alexander McCauley, Jonathon Shlens, and Dragomir Anguelov.
\newblock Large scale interactive motion forecasting for autonomous driving: The waymo open motion dataset.
\newblock In \emph{ICCV}, 2021{\natexlab{a}}.

\bibitem[Ettinger et~al.(2021{\natexlab{b}})Ettinger, Cheng, Caine, Liu, Zhao, Pradhan, Chai, Sapp, Qi, Zhou, et~al.]{ettinger2021large}
Scott Ettinger, Shuyang Cheng, Benjamin Caine, Chenxi Liu, Hang Zhao, Sabeek Pradhan, Yuning Chai, Ben Sapp, Charles~R Qi, Yin Zhou, et~al.
\newblock Large scale interactive motion forecasting for autonomous driving: The waymo open motion dataset.
\newblock In \emph{ICCV}, 2021{\natexlab{b}}.

\bibitem[Fadadu et~al.(2022)Fadadu, Pandey, Hegde, Shi, Chou, Djuric, and Vallespi-Gonzalez]{fadadu2022multi}
Sudeep Fadadu, Shreyash Pandey, Darshan Hegde, Yi Shi, Fang-Chieh Chou, Nemanja Djuric, and Carlos Vallespi-Gonzalez.
\newblock Multi-view fusion of sensor data for improved perception and prediction in autonomous driving.
\newblock In \emph{Proceedings of the IEEE/CVF Winter Conference on Applications of Computer Vision}, pages 2349--2357, 2022.

\bibitem[Gao et~al.(2020)Gao, Sun, Zhao, Shen, Anguelov, Li, and Schmid]{gao2020vectornet}
Jiyang Gao, Chen Sun, Hang Zhao, Yi Shen, Dragomir Anguelov, Congcong Li, and Cordelia Schmid.
\newblock Vectornet: Encoding hd maps and agent dynamics from vectorized representation.
\newblock In \emph{CVPR}, 2020.

\bibitem[Gilles et~al.(2021)Gilles, Sabatini, Tsishkou, Stanciulescu, and Moutarde]{gilles2021home}
Thomas Gilles, Stefano Sabatini, Dzmitry Tsishkou, Bogdan Stanciulescu, and Fabien Moutarde.
\newblock Home: Heatmap output for future motion estimation.
\newblock In \emph{2021 IEEE International Intelligent Transportation Systems Conference (ITSC)}, pages 500--507. IEEE, 2021.

\bibitem[Gilles et~al.(2022)Gilles, Sabatini, Tsishkou, Stanciulescu, and Moutarde]{gilles2022gohome}
Thomas Gilles, Stefano Sabatini, Dzmitry Tsishkou, Bogdan Stanciulescu, and Fabien Moutarde.
\newblock Gohome: Graph-oriented heatmap output for future motion estimation.
\newblock In \emph{2022 international conference on robotics and automation (ICRA)}, pages 9107--9114. IEEE, 2022.

\bibitem[Gu et~al.(2021)Gu, Sun, and Zhao]{DenseTNT_Gu_2021_ICCV}
Junru Gu, Chen Sun, and Hang Zhao.
\newblock Densetnt: End-to-end trajectory prediction from dense goal sets.
\newblock In \emph{ICCV}, 2021.

\bibitem[Gu et~al.(2023)Gu, Hu, Zhang, Chen, Wang, Wang, and Zhao]{gu2023vip3d}
Junru Gu, Chenxu Hu, Tianyuan Zhang, Xuanyao Chen, Yilun Wang, Yue Wang, and Hang Zhao.
\newblock Vip3d: End-to-end visual trajectory prediction via 3d agent queries.
\newblock In \emph{Proceedings of the IEEE/CVF Conference on Computer Vision and Pattern Recognition}, pages 5496--5506, 2023.

\bibitem[Gulino et~al.(2023)Gulino, Fu, Luo, Tucker, Bronstein, Lu, Harb, Pan, Wang, Chen, et~al.]{gulino2023waymax}
Cole Gulino, Justin Fu, Wenjie Luo, George Tucker, Eli Bronstein, Yiren Lu, Jean Harb, Xinlei Pan, Yan Wang, Xiangyu Chen, et~al.
\newblock Waymax: An accelerated, data-driven simulator for large-scale autonomous driving research.
\newblock \emph{arXiv preprint arXiv:2310.08710}, 2023.

\bibitem[Gupta et~al.(2018)Gupta, Johnson, Fei-Fei, Savarese, and Alahi]{gupta2018social}
Agrim Gupta, Justin Johnson, Li Fei-Fei, Silvio Savarese, and Alexandre Alahi.
\newblock Social gan: Socially acceptable trajectories with generative adversarial networks.
\newblock In \emph{Proceedings of the IEEE conference on computer vision and pattern recognition}, pages 2255--2264, 2018.

\bibitem[Hagedorn et~al.(2023)Hagedorn, Hallgarten, Stoll, and Condurache]{hagedorn2023rethinking}
Steffen Hagedorn, Marcel Hallgarten, Martin Stoll, and Alexandru Condurache.
\newblock Rethinking integration of prediction and planning in deep learning-based automated driving systems: A review.
\newblock \emph{arXiv preprint arXiv:2308.05731}, 2023.

\bibitem[Hong et~al.(2019)Hong, Sapp, and Philbin]{hong2019rules}
Joey Hong, Benjamin Sapp, and James Philbin.
\newblock Rules of the road: Predicting driving behavior with a convolutional model of semantic interactions.
\newblock In \emph{CVPR}, 2019.

\bibitem[Hu et~al.(2021)Hu, Murez, Mohan, Dudas, Hawke, Badrinarayanan, Cipolla, and Kendall]{hu2021fiery}
Anthony Hu, Zak Murez, Nikhil Mohan, Sof{\'\i}a Dudas, Jeffrey Hawke, Vijay Badrinarayanan, Roberto Cipolla, and Alex Kendall.
\newblock Fiery: Future instance prediction in bird's-eye view from surround monocular cameras.
\newblock In \emph{Proceedings of the IEEE/CVF International Conference on Computer Vision}, pages 15273--15282, 2021.

\bibitem[Hu et~al.(2022)Hu, Chen, Wu, Li, Yan, and Tao]{hu2022st}
Shengchao Hu, Li Chen, Penghao Wu, Hongyang Li, Junchi Yan, and Dacheng Tao.
\newblock St-p3: End-to-end vision-based autonomous driving via spatial-temporal feature learning.
\newblock In \emph{European Conference on Computer Vision}, pages 533--549. Springer, 2022.

\bibitem[Hu et~al.(2023)Hu, Yang, Chen, Li, Sima, Zhu, Chai, Du, Lin, Wang, et~al.]{hu2023planning}
Yihan Hu, Jiazhi Yang, Li Chen, Keyu Li, Chonghao Sima, Xizhou Zhu, Siqi Chai, Senyao Du, Tianwei Lin, Wenhai Wang, et~al.
\newblock Planning-oriented autonomous driving.
\newblock In \emph{Proceedings of the IEEE/CVF Conference on Computer Vision and Pattern Recognition}, pages 17853--17862, 2023.

\bibitem[Jia et~al.(2023{\natexlab{a}})Jia, Gao, Chen, Yan, Liu, and Li]{jia2023driveadapter}
Xiaosong Jia, Yulu Gao, Li Chen, Junchi Yan, Patrick~Langechuan Liu, and Hongyang Li.
\newblock Driveadapter: Breaking the coupling barrier of perception and planning in end-to-end autonomous driving.
\newblock In \emph{Proceedings of the IEEE/CVF International Conference on Computer Vision}, pages 7953--7963, 2023{\natexlab{a}}.

\bibitem[Jia et~al.(2023{\natexlab{b}})Jia, Wu, Chen, Liu, Li, and Yan]{jia2023hdgt}
Xiaosong Jia, Penghao Wu, Li Chen, Yu Liu, Hongyang Li, and Junchi Yan.
\newblock Hdgt: Heterogeneous driving graph transformer for multi-agent trajectory prediction via scene encoding.
\newblock \emph{IEEE transactions on pattern analysis and machine intelligence}, 2023{\natexlab{b}}.

\bibitem[Jia et~al.(2023{\natexlab{c}})Jia, Wu, Chen, Xie, He, Yan, and Li]{jia2023think}
Xiaosong Jia, Penghao Wu, Li Chen, Jiangwei Xie, Conghui He, Junchi Yan, and Hongyang Li.
\newblock Think twice before driving: Towards scalable decoders for end-to-end autonomous driving.
\newblock In \emph{Proceedings of the IEEE/CVF Conference on Computer Vision and Pattern Recognition}, pages 21983--21994, 2023{\natexlab{c}}.

\bibitem[Kamenev et~al.(2022)Kamenev, Wang, Bohan, Kulkarni, Kartal, Molchanov, Birchfield, Nist{\'e}r, and Smolyanskiy]{kamenev2022predictionnet}
Alexey Kamenev, Lirui Wang, Ollin~Boer Bohan, Ishwar Kulkarni, Bilal Kartal, Artem Molchanov, Stan Birchfield, David Nist{\'e}r, and Nikolai Smolyanskiy.
\newblock Predictionnet: Real-time joint probabilistic traffic prediction for planning, control, and simulation.
\newblock In \emph{2022 International Conference on Robotics and Automation (ICRA)}, pages 8936--8942. IEEE, 2022.

\bibitem[Khandelwal et~al.(2020)Khandelwal, Qi, Singh, Hartnett, and Ramanan]{khandelwal2020if}
Siddhesh Khandelwal, William Qi, Jagjeet Singh, Andrew Hartnett, and Deva Ramanan.
\newblock What-if motion prediction for autonomous driving.
\newblock \emph{arXiv preprint arXiv:2008.10587}, 2020.

\bibitem[Kirillov et~al.(2023)Kirillov, Mintun, Ravi, Mao, Rolland, Gustafson, Xiao, Whitehead, Berg, Lo, et~al.]{kirillov2023segment}
Alexander Kirillov, Eric Mintun, Nikhila Ravi, Hanzi Mao, Chloe Rolland, Laura Gustafson, Tete Xiao, Spencer Whitehead, Alexander~C Berg, Wan-Yen Lo, et~al.
\newblock Segment anything.
\newblock \emph{arXiv preprint arXiv:2304.02643}, 2023.

\bibitem[Konev et~al.(2021)Konev, Brodt, and Sanakoyeu]{Motioncnn2021}
Stepan Konev, Kirill Brodt, and Artsiom Sanakoyeu.
\newblock Motioncnn: A strong baseline for motion prediction in autonomous driving.
\newblock In \emph{CVPRW}, 2021.

\bibitem[Li et~al.(2022)Li, Wang, Li, Xie, Sima, Lu, Qiao, and Dai]{li2022bevformer}
Zhiqi Li, Wenhai Wang, Hongyang Li, Enze Xie, Chonghao Sima, Tong Lu, Yu Qiao, and Jifeng Dai.
\newblock Bevformer: Learning bird’s-eye-view representation from multi-camera images via spatiotemporal transformers.
\newblock In \emph{European conference on computer vision}, pages 1--18. Springer, 2022.

\bibitem[Liang et~al.(2020{\natexlab{a}})Liang, Yang, Hu, Chen, Liao, Feng, and Urtasun]{liang2020learning}
Ming Liang, Bin Yang, Rui Hu, Yun Chen, Renjie Liao, Song Feng, and Raquel Urtasun.
\newblock Learning lane graph representations for motion forecasting.
\newblock In \emph{ECCV}, pages 541--556. Springer, 2020{\natexlab{a}}.

\bibitem[Liang et~al.(2020{\natexlab{b}})Liang, Yang, Zeng, Chen, Hu, Casas, and Urtasun]{liang2020pnpnet}
Ming Liang, Bin Yang, Wenyuan Zeng, Yun Chen, Rui Hu, Sergio Casas, and Raquel Urtasun.
\newblock Pnpnet: End-to-end perception and prediction with tracking in the loop.
\newblock In \emph{CVPR}, 2020{\natexlab{b}}.

\bibitem[Liu et~al.(2021)Liu, Zeng, Urtasun, and Yumer]{liu2021deep}
Jerry Liu, Wenyuan Zeng, Raquel Urtasun, and Ersin Yumer.
\newblock Deep structured reactive planning.
\newblock In \emph{2021 IEEE International Conference on Robotics and Automation (ICRA)}, pages 4897--4904. IEEE, 2021.

\bibitem[Liu et~al.(2022)Liu, Zhou, Qi, Gong, Su, and Anguelov]{liu2022less}
Minghua Liu, Yin Zhou, Charles~R Qi, Boqing Gong, Hao Su, and Dragomir Anguelov.
\newblock Less: Label-efficient semantic segmentation for lidar point clouds.
\newblock In \emph{European conference on computer vision}, pages 70--89. Springer, 2022.

\bibitem[Loshchilov and Hutter(2017)]{Loshchilov2017DecoupledWD}
Ilya Loshchilov and Frank Hutter.
\newblock Decoupled weight decay regularization.
\newblock In \emph{International Conference on Learning Representations}, 2017.

\bibitem[Luo et~al.(2018)Luo, Yang, and Urtasun]{luo2018fast}
Wenjie Luo, Bin Yang, and Raquel Urtasun.
\newblock Fast and furious: Real time end-to-end 3d detection, tracking and motion forecasting with a single convolutional net.
\newblock In \emph{Proceedings of the IEEE conference on Computer Vision and Pattern Recognition}, pages 3569--3577, 2018.

\bibitem[Luo et~al.(2023)Luo, Park, Cornman, Sapp, and Anguelov]{luo2023jfp}
Wenjie Luo, Cheol Park, Andre Cornman, Benjamin Sapp, and Dragomir Anguelov.
\newblock Jfp: Joint future prediction with interactive multi-agent modeling for autonomous driving.
\newblock In \emph{Conference on Robot Learning}, pages 1457--1467. PMLR, 2023.

\bibitem[Mao et~al.(2023)Mao, Qian, Zhao, and Wang]{mao2023gpt}
Jiageng Mao, Yuxi Qian, Hang Zhao, and Yue Wang.
\newblock Gpt-driver: Learning to drive with gpt.
\newblock \emph{arXiv preprint arXiv:2310.01415}, 2023.

\bibitem[Marchetti et~al.(2020)Marchetti, Becattini, Seidenari, and Bimbo]{marchetti2020mantra}
Francesco Marchetti, Federico Becattini, Lorenzo Seidenari, and Alberto~Del Bimbo.
\newblock Mantra: Memory augmented networks for multiple trajectory prediction.
\newblock In \emph{Proceedings of the IEEE/CVF conference on computer vision and pattern recognition}, pages 7143--7152, 2020.

\bibitem[Nayakanti et~al.(2023)Nayakanti, Al-Rfou, Zhou, Goel, Refaat, and Sapp]{nayakanti2023wayformer}
Nigamaa Nayakanti, Rami Al-Rfou, Aurick Zhou, Kratarth Goel, Khaled~S Refaat, and Benjamin Sapp.
\newblock Wayformer: Motion forecasting via simple \& efficient attention networks.
\newblock In \emph{2023 IEEE International Conference on Robotics and Automation (ICRA)}, pages 2980--2987. IEEE, 2023.

\bibitem[Ngiam et~al.(2021)Ngiam, Caine, Vasudevan, Zhang, Chiang, Ling, Roelofs, Bewley, Liu, Venugopal, et~al.]{ngiam2021scene}
Jiquan Ngiam, Benjamin Caine, Vijay Vasudevan, Zhengdong Zhang, Hao-Tien~Lewis Chiang, Jeffrey Ling, Rebecca Roelofs, Alex Bewley, Chenxi Liu, Ashish Venugopal, et~al.
\newblock Scene transformer: A unified architecture for predicting multiple agent trajectories.
\newblock \emph{arXiv preprint arXiv:2106.08417}, 2021.

\bibitem[Ngiam et~al.(2022)Ngiam, Caine, Vasudevan, Zhang, Chiang, Ling, Roelofs, Bewley, Liu, Venugopal, Weiss, Sapp, Chen, and Shlens]{ngiam2022scenetransformer}
Jiquan Ngiam, Benjamin Caine, Vijay Vasudevan, Zhengdong Zhang, Hao-Tien~Lewis Chiang, Jeffrey Ling, Rebecca Roelofs, Alex Bewley, Chenxi Liu, Ashish Venugopal, David Weiss, Ben Sapp, Zhifeng Chen, and Jonathon Shlens.
\newblock Scene transformer: A unified architecture for predicting multiple agent trajectories.
\newblock In \emph{ICLR}, 2022.

\bibitem[Oquab et~al.(2023)Oquab, Darcet, Moutakanni, Vo, Szafraniec, Khalidov, Fernandez, Haziza, Massa, El-Nouby, et~al.]{oquab2023dinov2}
Maxime Oquab, Timoth{\'e}e Darcet, Th{\'e}o Moutakanni, Huy Vo, Marc Szafraniec, Vasil Khalidov, Pierre Fernandez, Daniel Haziza, Francisco Massa, Alaaeldin El-Nouby, et~al.
\newblock Dinov2: Learning robust visual features without supervision.
\newblock \emph{arXiv preprint arXiv:2304.07193}, 2023.

\bibitem[Park et~al.(2020)Park, Lee, Seo, Bhat, Kang, Francis, Jadhav, Liang, and Morency]{park2020diverse}
Seong~Hyeon Park, Gyubok Lee, Jimin Seo, Manoj Bhat, Minseok Kang, Jonathan Francis, Ashwin Jadhav, Paul~Pu Liang, and Louis-Philippe Morency.
\newblock Diverse and admissible trajectory forecasting through multimodal context understanding.
\newblock In \emph{ECCV}, pages 282--298. Springer, 2020.

\bibitem[Pomerleau(1988)]{pomerleau1988alvinn}
Dean~A Pomerleau.
\newblock Alvinn: An autonomous land vehicle in a neural network.
\newblock \emph{Advances in neural information processing systems}, 1, 1988.

\bibitem[Qi et~al.(2021)Qi, Zhou, Najibi, Sun, Vo, Deng, and Anguelov]{qi2021offboard}
Charles~R Qi, Yin Zhou, Mahyar Najibi, Pei Sun, Khoa Vo, Boyang Deng, and Dragomir Anguelov.
\newblock Offboard 3d object detection from point cloud sequences.
\newblock In \emph{CVPR}, 2021.

\bibitem[Radford et~al.(2021)Radford, Kim, Hallacy, Ramesh, Goh, Agarwal, Sastry, Askell, Mishkin, Clark, et~al.]{radford2021learning}
Alec Radford, Jong~Wook Kim, Chris Hallacy, Aditya Ramesh, Gabriel Goh, Sandhini Agarwal, Girish Sastry, Amanda Askell, Pamela Mishkin, Jack Clark, et~al.
\newblock Learning transferable visual models from natural language supervision.
\newblock In \emph{ICML}, 2021.

\bibitem[Rhinehart et~al.(2019)Rhinehart, McAllister, Kitani, and Levine]{rhinehart2019precog}
Nicholas Rhinehart, Rowan McAllister, Kris Kitani, and Sergey Levine.
\newblock Precog: Prediction conditioned on goals in visual multi-agent settings.
\newblock In \emph{Proceedings of the IEEE/CVF International Conference on Computer Vision}, pages 2821--2830, 2019.

\bibitem[Sadat et~al.(2020)Sadat, Casas, Ren, Wu, Dhawan, and Urtasun]{sadat2020perceive}
Abbas Sadat, Sergio Casas, Mengye Ren, Xinyu Wu, Pranaab Dhawan, and Raquel Urtasun.
\newblock Perceive, predict, and plan: Safe motion planning through interpretable semantic representations.
\newblock In \emph{ECCV}, pages 414--430. Springer, 2020.

\bibitem[Salzmann et~al.(2020)Salzmann, Ivanovic, Chakravarty, and Pavone]{salzmann2020trajectron++}
Tim Salzmann, Boris Ivanovic, Punarjay Chakravarty, and Marco Pavone.
\newblock Trajectron++: Dynamically-feasible trajectory forecasting with heterogeneous data.
\newblock In \emph{ECCV}, pages 683--700. Springer, 2020.

\bibitem[Seff et~al.(2023)Seff, Cera, Chen, Ng, Zhou, Nayakanti, Refaat, Al-Rfou, and Sapp]{seff2023motionlm}
Ari Seff, Brian Cera, Dian Chen, Mason Ng, Aurick Zhou, Nigamaa Nayakanti, Khaled~S Refaat, Rami Al-Rfou, and Benjamin Sapp.
\newblock Motionlm: Multi-agent motion forecasting as language modeling.
\newblock In \emph{Proceedings of the IEEE/CVF International Conference on Computer Vision}, pages 8579--8590, 2023.

\bibitem[Shi et~al.(2022)Shi, Jiang, Dai, and Schiele]{shi2022motion}
Shaoshuai Shi, Li Jiang, Dengxin Dai, and Bernt Schiele.
\newblock Motion transformer with global intention localization and local movement refinement.
\newblock \emph{Advances in Neural Information Processing Systems}, 35:\penalty0 6531--6543, 2022.

\bibitem[Song et~al.(2020)Song, Ding, Chen, Shen, Wang, and Chen]{song2020pip}
Haoran Song, Wenchao Ding, Yuxuan Chen, Shaojie Shen, Michael~Yu Wang, and Qifeng Chen.
\newblock Pip: Planning-informed trajectory prediction for autonomous driving.
\newblock In \emph{ECCV}, pages 598--614. Springer, 2020.

\bibitem[Sun et~al.(2022)Sun, Huang, Gu, Williams, and Zhao]{sun2022m2i}
Qiao Sun, Xin Huang, Junru Gu, Brian~C Williams, and Hang Zhao.
\newblock M2i: From factored marginal trajectory prediction to interactive prediction.
\newblock In \emph{Proceedings of the IEEE/CVF Conference on Computer Vision and Pattern Recognition}, pages 6543--6552, 2022.

\bibitem[Tang and Salakhutdinov(2019)]{tang2019multiple}
Charlie Tang and Russ~R Salakhutdinov.
\newblock Multiple futures prediction.
\newblock \emph{Advances in neural information processing systems}, 32, 2019.

\bibitem[Tolstaya et~al.(2021)Tolstaya, Mahjourian, Downey, Vadarajan, Sapp, and Anguelov]{tolstaya2021identifying}
Ekaterina Tolstaya, Reza Mahjourian, Carlton Downey, Balakrishnan Vadarajan, Benjamin Sapp, and Dragomir Anguelov.
\newblock Identifying driver interactions via conditional behavior prediction.
\newblock In \emph{2021 IEEE International Conference on Robotics and Automation (ICRA)}, pages 3473--3479. IEEE, 2021.

\bibitem[Varadarajan et~al.(2021)Varadarajan, Hefny, Srivastava, Refaat, Nayakanti, Cornman, Chen, Douillard, Lam, Anguelov, and Sapp]{MultiPathPP_2021}
Balakrishnan Varadarajan, Ahmed Hefny, Avikalp Srivastava, Khaled~S. Refaat, Nigamaa Nayakanti, Andre Cornman, Kan Chen, Bertrand Douillard, Chi{-}Pang Lam, Dragomir Anguelov, and Benjamin Sapp.
\newblock Multipath++: Efficient information fusion and trajectory aggregation for behavior prediction.
\newblock \emph{CoRR}, abs/2111.14973, 2021.

\bibitem[Varadarajan et~al.(2022)Varadarajan, Hefny, Srivastava, Refaat, Nayakanti, Cornman, Chen, Douillard, Lam, Anguelov, and Sapp]{multipath_plusplus2022}
Balakrishnan Varadarajan, Ahmed Hefny, Avikalp Srivastava, Khaled~S. Refaat, Nigamaa Nayakanti, Andre Cornman, Kan Chen, Bertrand Douillard, Chi~Pang Lam, Dragomir Anguelov, and Benjamin Sapp.
\newblock Multipath++: Efficient information fusion and trajectory aggregation for behavior prediction.
\newblock In \emph{2022 International Conference on Robotics and Automation (ICRA)}, pages 7814--7821, 2022.

\bibitem[Wang et~al.(2023)Wang, Zhu, Lu, Zhong, Chen, Shen, Wang, and Wang]{wang2023bevgpt}
Pengqin Wang, Meixin Zhu, Hongliang Lu, Hui Zhong, Xianda Chen, Shaojie Shen, Xuesong Wang, and Yinhai Wang.
\newblock Bevgpt: Generative pre-trained large model for autonomous driving prediction, decision-making, and planning.
\newblock \emph{arXiv preprint arXiv:2310.10357}, 2023.

\bibitem[Wilson et~al.(2023)Wilson, Qi, Agarwal, Lambert, Singh, Khandelwal, Pan, Kumar, Hartnett, Pontes, et~al.]{wilson2023argoverse}
Benjamin Wilson, William Qi, Tanmay Agarwal, John Lambert, Jagjeet Singh, Siddhesh Khandelwal, Bowen Pan, Ratnesh Kumar, Andrew Hartnett, Jhony~Kaesemodel Pontes, et~al.
\newblock Argoverse 2: Next generation datasets for self-driving perception and forecasting.
\newblock \emph{arXiv preprint arXiv:2301.00493}, 2023.

\bibitem[Xu et~al.(2023)Xu, Zhang, Xie, Zhao, Guo, Wong, Li, and Zhao]{xu2023drivegpt4}
Zhenhua Xu, Yujia Zhang, Enze Xie, Zhen Zhao, Yong Guo, Kenneth~KY Wong, Zhenguo Li, and Hengshuang Zhao.
\newblock Drivegpt4: Interpretable end-to-end autonomous driving via large language model.
\newblock \emph{arXiv preprint arXiv:2310.01412}, 2023.

\bibitem[Yu et~al.(2021)Yu, Li, Koh, Zhang, Pang, Qin, Ku, Xu, Baldridge, and Wu]{yu2021vector}
Jiahui Yu, Xin Li, Jing~Yu Koh, Han Zhang, Ruoming Pang, James Qin, Alexander Ku, Yuanzhong Xu, Jason Baldridge, and Yonghui Wu.
\newblock Vector-quantized image modeling with improved vqgan.
\newblock \emph{arXiv preprint arXiv:2110.04627}, 2021.

\bibitem[Zeng et~al.(2019)Zeng, Luo, Suo, Sadat, Yang, Casas, and Urtasun]{zeng2019end}
Wenyuan Zeng, Wenjie Luo, Simon Suo, Abbas Sadat, Bin Yang, Sergio Casas, and Raquel Urtasun.
\newblock End-to-end interpretable neural motion planner.
\newblock In \emph{CVPR}, 2019.

\bibitem[Zhang et~al.(2022{\natexlab{a}})Zhang, Chen, Wang, Wang, and Zhao]{zhang2022mutr3d}
Tianyuan Zhang, Xuanyao Chen, Yue Wang, Yilun Wang, and Hang Zhao.
\newblock Mutr3d: A multi-camera tracking framework via 3d-to-2d queries.
\newblock In \emph{Proceedings of the IEEE/CVF Conference on Computer Vision and Pattern Recognition}, pages 4537--4546, 2022{\natexlab{a}}.

\bibitem[Zhang et~al.(2022{\natexlab{b}})Zhang, Zhu, Zheng, Huang, Huang, Zhou, and Lu]{zhang2022beverse}
Yunpeng Zhang, Zheng Zhu, Wenzhao Zheng, Junjie Huang, Guan Huang, Jie Zhou, and Jiwen Lu.
\newblock Beverse: Unified perception and prediction in birds-eye-view for vision-centric autonomous driving.
\newblock \emph{arXiv preprint arXiv:2205.09743}, 2022{\natexlab{b}}.

\bibitem[Zhou and Tuzel(2018)]{zhou2018voxelnet}
Yin Zhou and Oncel Tuzel.
\newblock Voxelnet: End-to-end learning for point cloud based 3d object detection.
\newblock In \emph{CVPR}, 2018.

\bibitem[Zhou et~al.(2020)Zhou, Sun, Zhang, Anguelov, Gao, Ouyang, Guo, Ngiam, and Vasudevan]{zhou2020end}
Yin Zhou, Pei Sun, Yu Zhang, Dragomir Anguelov, Jiyang Gao, Tom Ouyang, James Guo, Jiquan Ngiam, and Vijay Vasudevan.
\newblock End-to-end multi-view fusion for 3d object detection in lidar point clouds.
\newblock In \emph{CoRL}, 2020.

\end{thebibliography}
}

\clearpage
\appendix
\section*{Appendix}

\section{Additional Qualitative Results} 
\begin{figure}
  \centering
  \includegraphics[width=\linewidth]{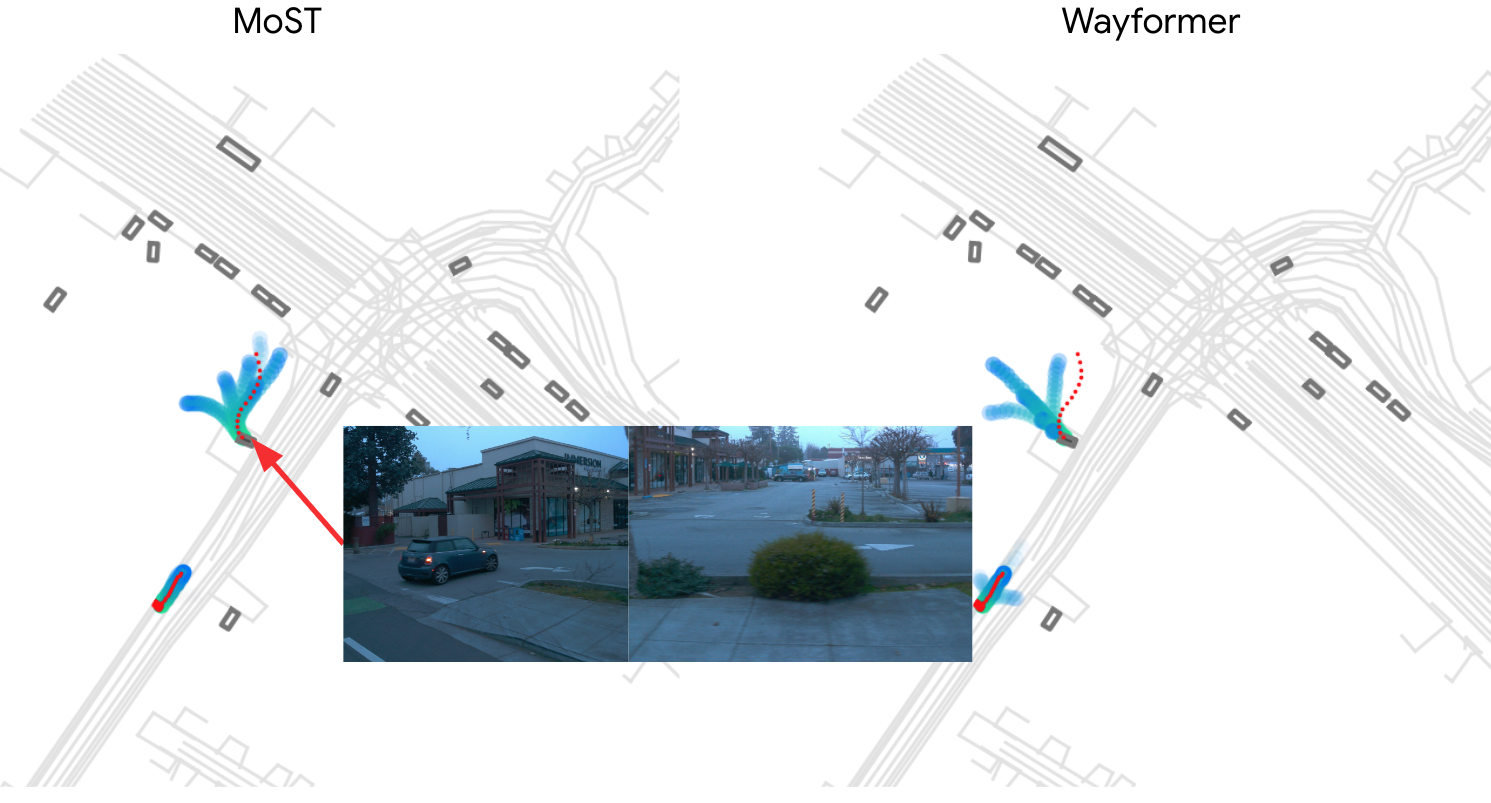}
  \caption{Additional qualitative comparison between \modelname{} and Wayformer~\cite{nayakanti2023wayformer} baseline. The agent boxes are colored
by their types: gray for vehicle, red for pedestrian, and cyan for
cyclist. The predicted trajectories are ordered temporally from green (+0s) to blue (+8.0s). For each modeled agent, the models predict 6 trajectory candidates, whose confidence scores are illustrated by transparency: the more confident, the more visible. Ground truth trajectory is shown as red dots. Note that the vehicle indicated by the red arrow is entering a plaza which has no map coverage. Since our model has access to the rich visual signals, it correctly predicts the vehicle's possible trajectory which includes follows the arrow and turn right. Wayformer, on the other hand, completely missed this possibility due to the lack of road graph information in that region. } 
  \label{fig:supp_qualitative}
\end{figure}

An additional qualitative comparison can be found in Figure~\ref{fig:supp_qualitative}. In this scenario, the model is asked to predict the future trajectory of a vehicle entering a plaza which is not mapped by the road graph. Our model with access to visual information correctly predicts several trajectories following the arrow painted on the ground and turning right. 

\section{WOMD Camera Embeddings}

\paragraph{VQGAN Embedding} To extract VQGAN embedding for an image, we first resize the image into shape of $256\times 512$. Then we horizontally split the image into two patches and apply pre-trained ViT-VQGAN~\cite{yu2021vector} model on each patch respectively. Each patch contains $16\times 16$ tokens so each camera image can be represented as 512 tokens. The code-book size is 8192. 

\begin{figure}
  \centering
  \includegraphics[width=0.9\linewidth]{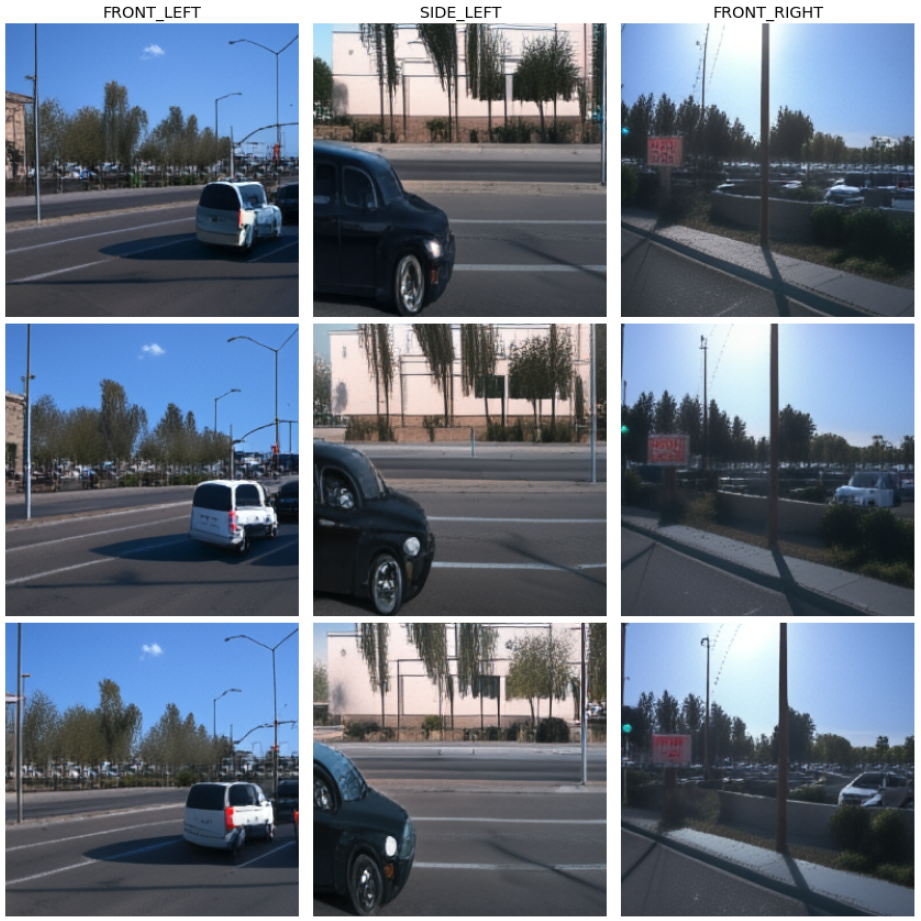}
  \caption{Examples of reconstructed driving images from ViT-VQGAN codes. We show 3 cameras at 3 consecutive timestamps. We are able to decode high quality images from VQGAN codes.}
  \label{fig:vqgan_decode}
\end{figure}

\begin{figure}
  \centering
  \includegraphics[width=\linewidth]{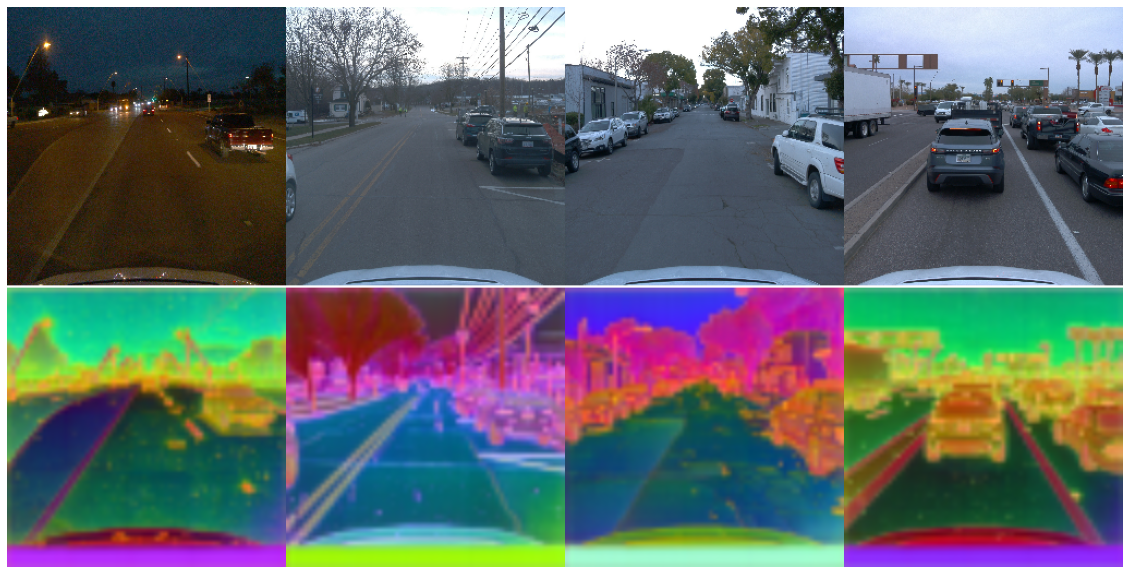}
  \caption{Examples of SAM feature. The first row shows camera images and the second row illustrates the SAM feature map visualized by PCA reduction from 256 to 3 dimensions.}
  \label{fig:sam_vis}
\end{figure}

\paragraph{SAM-H Embedding} 
For each camera we extract SAM ViT-H~\cite{kirillov2023segment} embedding of size $64\times 64\times 256$. Compared to VQGAN embeddings, SAM features are less spatially compressed due to its high-resolution feature map. The visualization of SAM Embedding can be found in Figure~\ref{fig:sam_vis}. We release the SAM features pooled per-scene-element.

\section{Implementation Details}

\paragraph{Model Detail} 
We use $N_{\text{elem}}^{\text{agent}} = 128$, $N_{\text{elem}}^{\text{open-set}} = 384$, $N_{\text{elem}}^{\text{gnd}} = 256$, and $N_{\text{pts}} = 65536$ in our experiments. We use sensor data from past 10 frames that correspond to the 1 second history and the current frame (i.e. $T=11$). Following Wayformer~\cite{nayakanti2023wayformer}, we train our model to output $K$ modes for the Gaussian mixture, where we experiment with $K=\{6,64\}$. During inference, we draw $2048$ samples from the predicted Gaussian mixture distribution, and use K-Means clustering to aggregate those 2048 samples into 6 final trajectory predictions. 

\paragraph{Training Detail} 
For all experiments, we train our model using AdamW~\cite{Loshchilov2017DecoupledWD} on 64 Google Cloud TPUv4 cores\footnote{\url{https://cloud.google.com/tpu}} with a global batch size of 512. We use a cosine learning rate schedule, where the learning rate is initialized to $3 \times 10^{-4}$ and ramps up to $6 \times 10^{-4}$ after 1,000 steps. The training finishes after 500,000 steps.

\paragraph{Notations} Please refer to Table~\ref{tbl:notation} for a summary of the notations used in the main paper. 

\begin{table}
    \setlength\tabcolsep{0.5pt}
    \centering
    \small
    \begin{tabular}{c|cc}
        \toprule
        Variable name & Description & Tensor Shape \\
        \midrule
        $\mathbf{N}_{\text{pts}}$ & \begin{tabular}[c]{@{}c@{}}The total number of  \\LiDAR points after down-sampling. \end{tabular}  & 1  \\
        $\mathbf{N}_{\text{elem}}$ & \begin{tabular}[c]{@{}c@{}}The total number of  scene elements. \end{tabular}  & 1  \\
        $T$ & \begin{tabular}[c]{@{}c@{}}The total number of  frames. \end{tabular}  & 1  \\
        $D$ & \begin{tabular}[c]{@{}c@{}}The feature dimension. \end{tabular}  & 1  \\
        \hline
        $\mathbf{P}_{\text{xyz}}$ & \begin{tabular}[c]{@{}c@{}}The aggregated \\LiDAR points from all frames \\after downsampling \end{tabular}  & $N_{\text{pts}}\times 3$  \\
        $\mathbf{P}_{\text{ind}}$ & \begin{tabular}[c]{@{}c@{}}The scene element index \\ and frame index for each LiDAR point \end{tabular}  & $N_{\text{pts}}\times 2$  \\
        $\mathbf{F}_{\text{pts}}$ & \begin{tabular}[c]{@{}c@{}}The per point image feature.\end{tabular}  & $N_{\text{pts}}\times D$  \\
        $\mathbf{B}$ & \begin{tabular}[l]{@{}l@{}}The box attributes, including box\\ center, box size, and box heading.\end{tabular}  & $N_{\text{elem}}\times T \times 7$  \\
        \hline
        $\mathbf{F}_{\text{img}}$ & \begin{tabular}[c]{@{}c@{}}The per scene- \\ element image feature.\end{tabular}  & $N_{\text{elem}}\times T\times D$  \\
        $\mathbf{F}_{\text{geo}}$ & \begin{tabular}[c]{@{}c@{}}The per scene-\\ element geometry feature.\end{tabular}  & $N_{\text{elem}}\times T\times D$  \\
        $\mathbf{f}_{\text{temporal}}$ & \begin{tabular}[c]{@{}c@{}}The learnable temporal embedding.\end{tabular}  & $1\times T \times D$  \\
        \bottomrule
    \end{tabular}
    \caption{Descriptions for variables used in the main paper. }
    \label{tbl:notation}
\end{table}

\end{document}